\documentclass[journal]{IEEEtran} 
\usepackage{cite}
\usepackage{amsmath,amssymb,amsfonts}
\usepackage{graphicx}
\usepackage{textcomp}
\usepackage[usenames, dvipsnames]{color}
\usepackage{booktabs}
\usepackage{array} 
\usepackage{ulem}
 \usepackage{multirow}
\usepackage{algorithm}
\usepackage{algpseudocode} 
\usepackage{tabularx,threeparttable}

 \newcommand{\hz}[1]{\textcolor{black}{#1}}

\begin{document}
\title{{DLTTA: Dynamic Learning Rate for Test-time Adaptation on Cross-domain Medical Images}}

\author{Hongzheng Yang, Cheng Chen,~\IEEEmembership{Member,~IEEE}, Meirui Jiang,~\IEEEmembership{Student Member,~IEEE}, \\ Quande Liu,~\IEEEmembership{Student Member,~IEEE}, Jianfeng Cao, Pheng Ann Heng,~\IEEEmembership{Senior \\ Member,~IEEE}, Qi Dou,~\IEEEmembership{Member,~IEEE}
\thanks{Hongzheng Yang is with the Department of Artificial Intelligence, Beihang University, Beijing, China. This work was done when H.Z. Yang did internship with CUHK.
Cheng Chen, Meirui Jiang, Quande Liu, Jianfeng Cao, Pheng Ann Heng and Qi Dou are with the Department of Computer Science and Engineering, The Chinese University of Hong Kong, Hong Kong, China. \\
Corresponding author: Cheng Chen (cchen@cse.cuhk.edu.hk).}
}
\maketitle

\begin{abstract}
Test-time adaptation (TTA) has increasingly been an important topic to efficiently tackle the cross-domain distribution shift at test time for medical images from different institutions. 
\hz{Previous TTA methods have a common limitation of using a fixed learning rate for all the test samples. Such a practice would be sub-optimal for TTA, because test data may arrive sequentially therefore \hz{the scale of distribution shift would change frequently}.}
\hz{To address this problem, we propose a novel dynamic learning rate adjustment method for test-time adaptation, called \textit{DLTTA}, which dynamically modulates the amount of weights update for each test image to account for the differences in their distribution shift.}
Specifically, our DLTTA is equipped with a memory bank based estimation scheme to effectively measure the discrepancy of a given test sample. 
Based on this estimated discrepancy, a dynamic learning rate adjustment strategy is then developed to achieve a suitable degree of adaptation for each test sample.
The effectiveness and general applicability of our DLTTA is extensively demonstrated on \hz{three tasks including retinal optical coherence tomography (OCT) segmentation, histopathological image classification, and prostate 3D MRI segmentation.}
Our method achieves effective and fast test-time adaptation with consistent performance improvement over current state-of-the-art test-time adaptation methods. Code is available at: https://github.com/med-air/DLTTA.

\end{abstract}


\section{Introduction}
\label{sec:introduction}
Despite recent progress on domain adaptation techniques~\cite{zhao2020review,kamnitsas2017unsupervised,zhang2018task,chen2020unsupervised}, 
deep learning models remain difficult to generalize across medical datasets with heterogeneous data distributions, that are caused by varying image acquisition conditions (e.g., imaging protocols and scanners)~\cite{ghafoorian2017transfer,gibson2018inter}.
This problem severely hinders the deployment of established deep learning models to new test samples with unknown data distributions.
Existing unsupervised domain adaptation (UDA) methods~\cite{huo2018synseg,DouOCCH18,zhang2018translating,varsavsky2020test, zhang2018multi,zhao2018supervised,bateson2020source,chen2021source, tzeng2017adversarial}
typically require to assemble a large test dataset to use concurrently with the training data to conduct distribution alignment for improved prediction performance on test samples. 
However, this setting is still problematic for real-world model deployment due to two important reasons.
Firstly, it is not efficient, if realistic, to wait for the accumulation of sufficient amount of test samples (e.g., thousands of instances), since the test data usually arrive sequentially one by one, or batch by batch. However, immediate prediction on a single test sample or a batch of instances is \hz{highly desired} in clinical practice for timely diagnosis and treatment for patients.
Secondly, accessing the training dataset at test time is practically difficult, because data sharing across hospitals is prohibitive due to the privacy concern for medical data.
Regarding these limitations, it would be more convenient if a given trained model could be quickly and continuously adapted to each test sample at inference time, without using any training data.
	\begin{figure}
	
		\centering
		\includegraphics[width=0.5\textwidth]{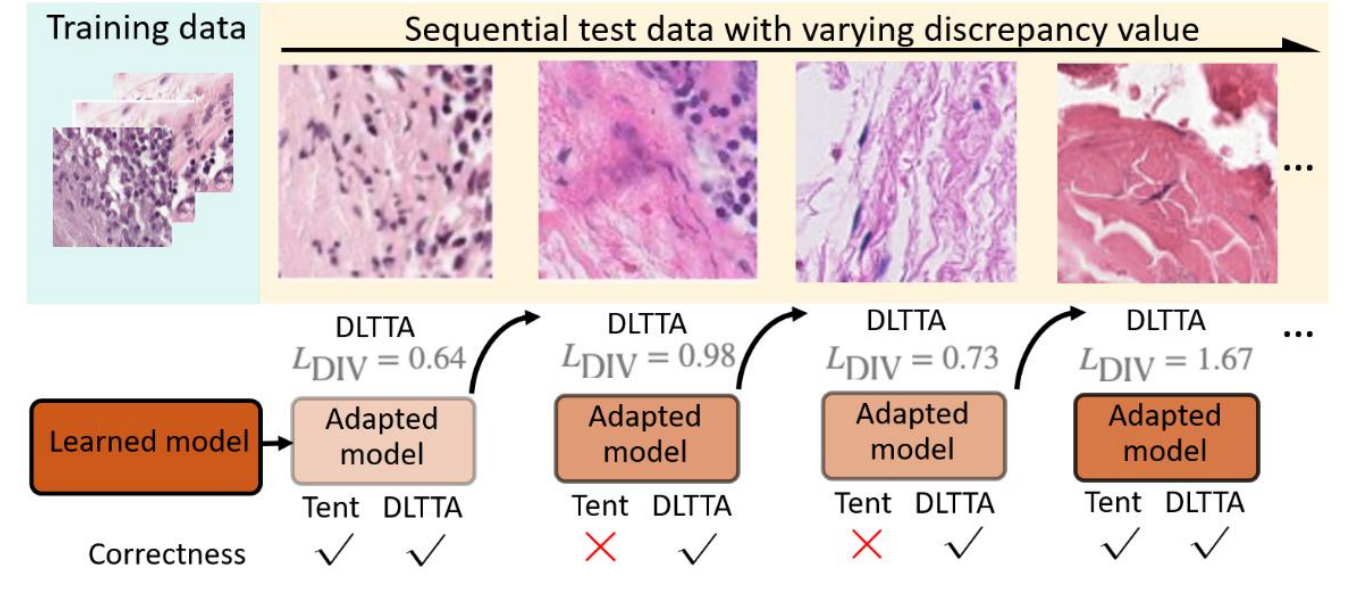}
		\vspace{-2mm}
		\caption{Illustration of the varying distribution shift of test data and \hz{our dynamic learning rate adjustment strategy for test-time adaptation (DLTTA)}. 
		Test images even from the same dataset present large appearance variations compared to the training data. Our estimated discrepancy values $L_\text{DIV}$ also indicate the distribution shift variations.
		By adapting the model dynamically, our DLTTA obtains more accurate predictions than the state-of-the-art test-time adaptation method Tent.
		}	
		\vspace{-0mm}
		\label{fig:cover}
	\end{figure}

This motivates the topic of test-time adaptation (TTA), and a few promising early investigations have been very recently conducted~\cite{sun2020test,wang2020tent,nado2020evaluating,he2020sdan,karani2021test}. The general idea is to gradually adjust model parameters by exploiting the distributional information provided by each test sample.
A representative work is test-time training (TTT)~\cite{sun2020test} which adds an auxiliary branch with self-supervision of rotation prediction to adapt the model encoder with the test data. 
Sharing similar idea, the method DTTA~\cite{karani2021test} and ATTA~\cite{he2021autoencoder} employ autoencoders to learn shape priors and to align feature space respectively, achieving promising results on cross-domain medical image analysis at test time. 
The latest state-of-the-art fully test-time adaptation method is Tent~\cite{wang2020tent} which proposes to adapt the batch normalization (BN) layer by minimizing the entropy of model predictions on test data. 

In this paper, we identify the common limitation in current test-time adaptation literature, i.e., applying the same learning rate for all the test samples.
This existing practice is sub-optimal since the sequentially arriving test data may not have the same distribution shift, hence can differ largely in their adaptation demand. For example, as shown in Fig.~\ref{fig:cover}, the test samples can present apparently varying degree of appearance changes, even though they come from the same test dataset (e.g., images stained in one hospital). This matters for test-time adaptation of the learned models.
Intuitively, for test data with severe distribution shift, the model requires greater update; while for test data with mild distribution shift in comparison to training statistics, the model should not be adapted too dramatically, otherwise might drift away from the discriminative features learned from the massive training data. 
Moreover, for test-time adaptation, the model is continuously updated. Therefore, even for two test samples with similar distribution shift but arrive at different time points, the demanded degree of adaptation would also be different due to the transitions in model parameters over time. 
\hz{To achieve such an adaptation demand, we argue it is important to dynamically adjust the test-time learning rate for each test sample, as learning rate is core to control the scale of updating the model weights in response to the estimated TTA loss during optimization.}

With above insights, \hz{we propose a novel dynamic learning rate method for test-time adaptation \textit{(DLTTA)}}.
\hz{The goal is to adapt the model dynamically to account for the differences in adaptation demand across test samples that arrive sequentially.} Firstly, to capture the progressive change of the model, we maintain a memory bank to cache the feature and prediction pairs of previous test data in a dynamic manner. 
Then, to estimate the discrepancy of a newly coming test sample, we retrieve semantically similar ones from the memory bank and calculate the Kullback–Leibler (KL) divergence between the cached predictions and the current output for the derived discrepancy. 
Based on the estimated discrepancy, we further adjust the learning rate to achieve suitable adaptation for online test data.
Compared with previous methods, the dependence on the choice of initial learning rate can be substantially alleviated owing to our adaptive learning rate strategy. 
Importantly, our method is designed to be simple yet effective, so that it can be flexibly incorporated to different existing TTA frameworks to improve adaptation performance, as well as be generally applied to different medical image analysis tasks with various model architectures.  
Our main contributions are as follows:
\begin{itemize}
\item We propose a new dynamic learning rate adjustment framework for test-time adaptation on cross-domain medical image analysis.
To the best of our knowledge, this is the first work of its kind to explore a dynamic learning strategy to overcome the varying distribution shift of inference data for model adaptation at test time.

\item We devise a novel memory bank-based discrepancy estimation strategy to model the adaptation demand, based on which the learning rate is \hz{specifically} adjusted for each particular test sample in an online manner.

\item We have validated the effectiveness of our method for both classification and segmentation tasks on \hz{three different medical imaging modalities with 2D or 3D models}. Experimental results show that our DLTTA is generic to different network architectures and can consistently outperform current state-of-the-art TTA methods. 

\end{itemize}

\begin{figure*}[!t]
	\centering
	\vspace{-1mm}
	\includegraphics[width=0.97\textwidth]{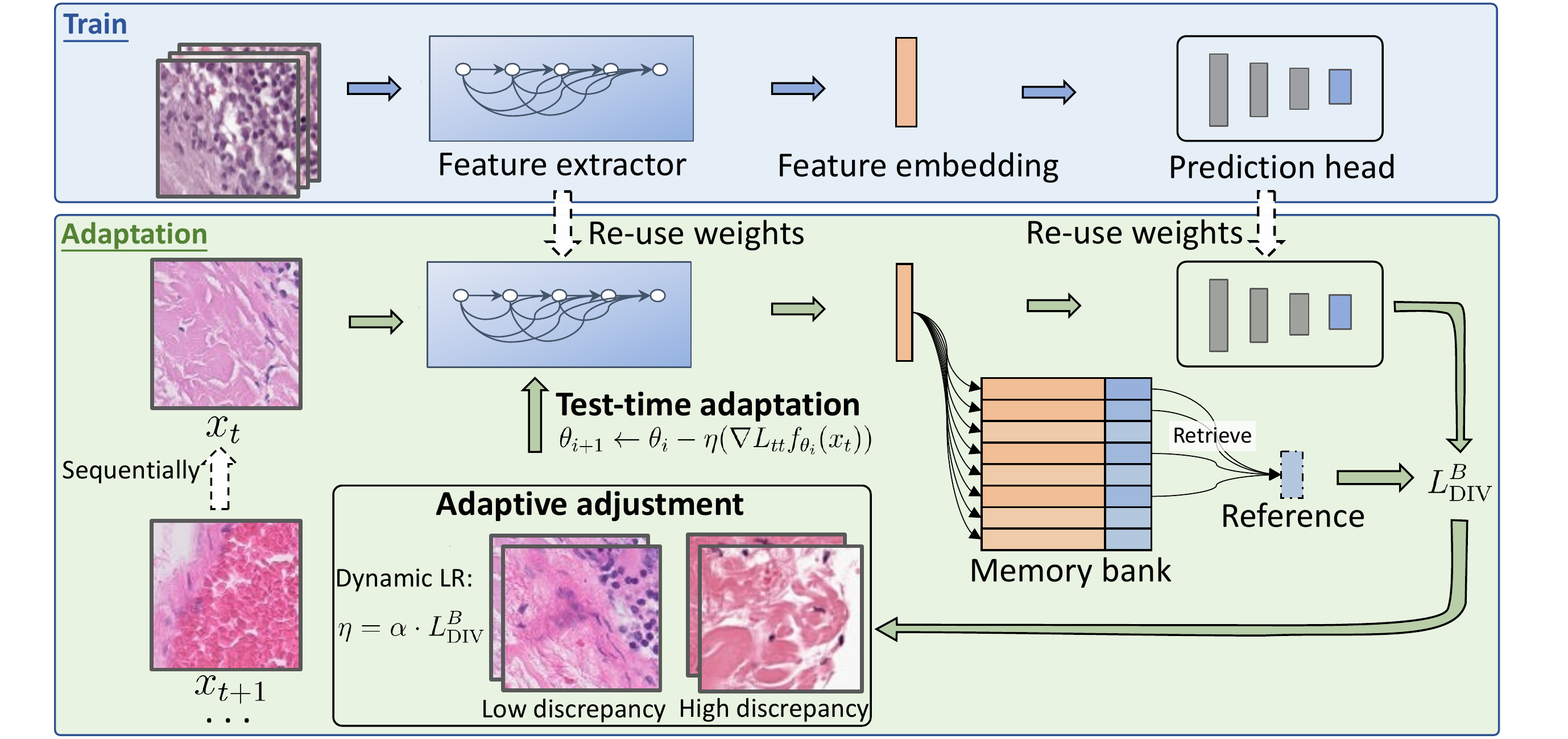}
	\vspace{0mm}
	\caption{The overview of our novel method for dynamic learning rate adjustment of test-time adaptation (DLTTA), to deal with the varying distribution shift of sequentially coming test data. A memory bank is constructed to derive the discrepancy $L_\text{DIV}^B$ between the current predictions and the semantically similar reference retrieved and calculated from the memory bank, and the test-time learning rate is then adaptively adjusted based on the estimated discrepancy to achieve desired adaptation for each test sample.}
	\label{fig:method}
\end{figure*}
\section{related works}

\subsection{Test-time Adaptation}
Adapting deep models solely based on unlabeled test data that arrive sequentially has recently drawn increasing interests. 
Unlike conventional UDA methods that require to access the training data and assemble sufficient amount of test data, the test-time adaptation methods are able to update a model with the distributional information \hz{provided by a single or a batch of test data}~\cite{sun2020test,wang2020tent,nado2020evaluating,liu2021ttt++,iwasawa2021prototyp,pandey2021generalization}. 
Prediction-time batch normalization (PTBN)~\cite{nado2020evaluating} re-estimates BN statistics~\cite{ioffe2015batch} based on the test data, which is simple but effective. 
TTT~\cite{sun2020test} adapts the feature extractor at test time by leveraging an auxiliary self-supervised task of rotation prediction.
TTT++~\cite{liu2021ttt++} improves TTT by further aligning the first- and second-order statistics of the training and test data and adopts the self-supervised task of contrastive learning~\cite{chen2020simclr}. 
Tent~\cite{wang2020tent} proposes to adapt the affine parameters in BN layers at test time by minimizing entropy of model predictions.
T3A~\cite{iwasawa2021prototyp} adjusts the classifier of a trained source model by computing a pseudo-prototype representation of different classes using unlabeled test data.

Tackling cross-domain distribution shift under more realistic settings has also been recently attempted in medical image applications~\cite{bateson2020source,varsavsky2020testUDA,chen2021source,liu2021adapting,he2020sdan,karani2021test,he2021autoencoder,zhu2021regis}. 
Test-time UDA~\cite{varsavsky2020testUDA} performs adversarial learning on each test sample separately to align the distribution with the source training data. 
Liu et al.~\cite{liu2021generative} propose a generative self-training test-time UDA framework for cross-domain magnetic resonance imaging synthesis.
Zhu et al.~\cite{zhu2021regis} design test-time training on each test image pair to improve the generalization of learning-based registration.
\hz{Wang et al.~\cite{wang2018interactive} make a model adaptive to a specific test image by bounding box and scribble-based interactive fine-tuning.}
To achieve test-time adaptation, DTTA~\cite{karani2021test} trains denoising autoencoders in the source domain to learn shape priors for adaptation at test time.
Similarly, Valvano et al.~\cite{valvano2021stop} keep mask discriminators to provide shape prior and fine-tune the segmentor on each individual test instance. 
The latest method ATTA~\cite{he2021autoencoder} trains a set of autoencoders on the source dataset and updates a set of adaptors at test time with autoencoders' reconstruction loss, showing more efficient inference and better results than previous works.

\hz{Existing test-time adaptation methods mainly focus on designing different TTA losses, but adopt the same step size for adaptation on all test samples. 
The strategy of a fixed step size is sub-optimal since for real-world data streams, the test data distributions may change frequently in a non-stationary way~\cite{venkataramani2018continual}.
When adapting a model to new environment at test time, we consider dynamic learning rate adjustment is important to control the amount of weights update at each iteration for effective and efficient model adaptation.}

\subsection{Dynamic Learning Rate}
\hz{To adjust the learning rate during a model training process, some predefined learning rate schedulers, such as linear step decay~\cite{goyal2017step_decay}, exponential decay, cosine/sine annealing~\cite{loshchilov2016anneal}, have been proposed and widely used in training deep neural networks. 
However, naively decaying or cycling the learning rate might be insufficient for complex situations, such as the test-time adaptation when facing the new environment with varying distribution shift. 
There are also works that exploit gradients for learning rate adjusting. Maclaurin et al.~\cite{maclaurin2015hyper_gradient} introduce a reversible learning technique to compute gradients with respect to the learning hyperparameters. The learning rate is then adjusted through an inner optimization. Baydin et al.~\cite{baydin2018online_hypergradient} compute the gradient with respect to the learning rate and dynamically adjusts the learning rate updates in an online manner at each iteration.  Although these methods can dynamically adjust the learning rate by gradient update, accurate supervision from labeled data is needed to calculate the update gradient signals, which is not available in test-time adaptation scenarios.}

\hz{Our insight is to achieve dynamic learning rate adjustment according to a estimated distribution shift. How to measure the discrepancy solely based on the model parameters and the current test sample in an unsupervised way is challenging and remains unsolved yet.}
A recent work~\cite{dda} which requires source training data for adaptation proposes to infer ``easier" and ``harder" test data by calculating a confidence score based on the output consistency of multiple classifiers. 
Although this work shares similar idea to ours on estimating the discrepancy of test images, it restricts to specific network architectures with multiple different classifiers and it uses the identified ``easier" test data to produce pseudo labels instead of designing a dynamic adaptation strategy. 
Lee et al.~\cite{lee2021unsupervised} exploit the stored BN statistics of training data to compare with the test data as a measure of distribution shift. Due to the continuous update of the model parameters at test time, the BN statistics of training data fail to represent the updated model status, leading to inaccurate discrepancy estimation. 
Our proposed memory bank-based discrepancy measurement captures both the model progress and the distribution variation of test data, thus can provide more up-to-date discrepancy estimation for effective dynamic learning rate adjustment of test-time adaptation.

\section{Methods}

In this section, we first provide the overall framework of test-time adaptation.
We then introduce our method for dynamic learning rate of test-time adaptation, in which a memory bank is incorporated to estimate the discrepancy of predictions at inference stage. With the estimated discrepancy, a dynamic learning rate is devised to dynamically modulate the test-time adaptation process \hz{for} stable and improved adaptation results. An overview of our framework for effective test-time adaptation is illustrated in Fig.~\ref{fig:method}.

\subsection{Test-time Adaptation Overall Framework}
When applying a model to new test data, a problem is the new test samples may follow an unknown data distribution, leading to severe performance drop of model prediction. Previous domain adaptation methods require to assemble a sufficient number of test data, while in real-world applications, the test sample usually arrives sequentially with varying distribution shift. The more appealing solution, test-time adaptation, aims to continuously adapt the model directly according to each presented test sample. 
Specifically, for test-time adaptation, we are only given a pre-trained model $f_{\theta^{s}}$ parameterized by a set of parameters $\theta^s$ and new test samples $\{x^{}_1,x^{}_2,...,x^{}_{t-1},x^{}_{t}\}$ that arrive online sequentially with varying distribution shift. The model $f_{\theta^{s}}$ is optimized with source training data $\{(x^{s}_i,y^{s}_i)\}_{i=1}^{N^{s}}$ by empirical risk minimization as below:
\begin{equation}
   \theta^{s}=\arg\underset{\theta}{\min} \frac{1}{N^s} \sum_{i=1}^{N{^{s}}}L_{m}(f_{\theta}(x_i^{s}),y_i^{s}),
    \label{eq:train time objective}
\end{equation}
where $L_{m}$ denotes the supervised loss for training, such as the cross entropy loss. 
The model $f_{\theta^{s}}$ would perform poorly on new test sample $x^{}_{t}$ that arrives at time step $t$ and differs from the source training samples in data distributions. 
Then to continuously adapt the model to achieve better generalization performance, a test-time objective function $L_{tt}$ needs to be dedicatedly designed to update the model parameters based on the distributional information provided by each test sample.
For a test sample $x_t$ appearing at the $t$-th iteration of \hz{model update}, we have:
\begin{equation}
    \theta_{t+1} \gets \theta_{t} - \eta ( \nabla L_{tt}f_{\theta_t}(x_t)),
    \label{eq:update}
\end{equation}
where $\eta$ denotes the learning rate for test-time adaptation, $\theta_1$ is initialized with $\theta^s$. The updated model $f_{\theta_{t+1}}$ is used to obtain prediction of test sample $x_t$.
Different test-time objective function $L_{tt}$ has been proposed in prior works, such as \hz{rotation prediction loss~\cite{sun2020test}, entropy minimization of model predictions~\cite{wang2020tent}, and autoencoder reconstruction loss~\cite{he2021autoencoder}}.

\vspace{0mm}
\subsection{Dynamic Learning Rate on Test Data}
\vspace{0mm}
Previous test-time adaptation methods adopt a fixed learning rate $\eta$ in Eq.~(\ref{eq:update}), which needs to be carefully chosen since the adaptation performance is sensitive to the learning rate~\cite{wang2020tent}. We argue that a static learning rate for all adaptation steps cannot accurately update the model to overcome the varying distribution shift of test data. 
We therefore propose a dynamic learning strategy to capture both the model changes and different shift degree of test data during test-time adaptation. 

\subsubsection{Memory Bank Construction for Discrepancy Estimation}
For dynamic adaptation, it is critical to know the demanded adaptation extent at each update step, which is the key factor of adjusting the learning rate. 
\hz{To measure the extent of adaptation, we propose a memory bank-based discrepancy estimation.}
The memory bank is constructed to store the latest pairs of feature representation and prediction mask extracted by the continuously updated model. Such pairs reflect the \hz{model's change}, and can be further utilized to calculate the distance with incoming test samples to estimate the distribution shift degree. With the shift degree measurement, we are able to adjust the test-time learning rate accordingly.

Specifically, the memory bank $\mathcal{M}$ comprises of $K$ pairs of keys and values \hz{$\{(q_k,v_k)\}_{k=1}^K$}. As shown in Fig.~\ref{fig:method}, the keys \hz{$\{q_k|q_k=h(x_t)\}$} are feature maps computed by the feature extractor $h$ of a model and the values $\{v_k|v_k=g(q_k)\}$ correspond to the prediction masks generated by the classifier head $g$. We update $\mathcal{M}$ by caching new $(q_k,v_k)$ pairs continuously. 
Since the model is progressively updated at test time, the early elements in the memory bank cannot indicate the most recent model performance on test data. We thus only maintain the memory bank with a fixed size $K$ and holds the First In First Out (FIFO) principle when writing new pairs. 
Then to measure the discrepancy of a newly coming test sample $x_t$, we aim to retrieve a support set $\mathcal{R}=\{(q_d,v_d)\}_{d=1}^D\subseteq{\mathcal{M}}$ from the memory bank, containing features and predictions of elements that are semantically similar to the current test sample $x_t$.
This is achieved by computing the $D$-nearest neighbors based on the $\text{L}2$ distance between each key and the features of the query sample $h(x_t)$. 
Since the key maintains semantic-level contextual features, the support set $\mathcal{R}$ retains internal agreement regarding high-level information, e.g., object categories in an image. The ensembling of predictions in $\mathcal{R}$ can be used as the reference prediction for $x_t$ as: 
\vspace{0mm}
\begin{equation}
     \hat{p}_t  = \frac{1}{D} \sum_{d=1}^{D} v_d.
    \label{eq:p_t}
\end{equation}
Then the prediction discrepancy of image $x_t$ is derived by:
	\begin{equation}
\begin{aligned}
     L_\text{DIV}  &= \frac{1}{2} (L_\text{KL}(\hat{p}_t||f_{\theta_t}(x_t)) + L_\text{KL}(f_{\theta_{t}}(x_t)||\hat{p}_{t})),\\ 
     &~\textnormal{with} ~ L_\text{KL}(\hat{p}_t||f_{\theta_t}(x_t)) = \sum_{j} \hat{p}_{t(j)} \textnormal{log} \frac{\hat{p}_{t(j)}}{f_{\theta_t}(x_{t(j)})},
     \label{eq:KL}
\end{aligned}
\end{equation}
where $L_\text{KL}$ denotes the KL divergence and $f_{\theta_t}(x_t)$ denotes the prediction of $x_t$ generated by the model $f_{\theta_t}$.

\subsubsection{Adaptive Learning Rate Adjustment}
With the discrepancy measured with Eq.~(\ref{eq:KL}), our intuition is that high discrepancy indicates a significant gap that needs to be largely bridged and low discrepancy requires smaller adaptation.
We therefore propose to dynamically adjust the self-supervised learning rate based on the estimated discrepancy. 
\hz{To describe our method in a more general way, we adopt a batch-wise formulation to account for that the test samples may arrive one by one or batch by batch. A single test image corresponds to the batch size 1.}
Given a batch of test data $\{x_{t,b}\}_{b=1}^{B}$ at each test-time adaptation step, we calculate the total discrepancy by:
\begin{equation}
     L_\text{DIV}^B  = \frac{1}{B} \sum_{b=1}^{B} L_{\text{DIV},b},   
    \label{eq:batchKL}
\end{equation}
where $B$ is the batch size. $L_\text{DIV}^B$ captures the general discrepancy distributed over batched predictions. Thereafter, batch-wise dynamic learning rate is obtained through a function $H(L_\text{DIV}^B)$ which outputs learning rate for test-time adaptation task directly based on the integral discrepancy $L_\text{DIV}^B$:
\begin{equation}
     H(L_\text{DIV}^B) =\hz{\alpha \cdot L_\text{DIV}^B}, 
    \label{eq:h_div}
\end{equation}
where $\alpha$ scales the learning rate further and could be empirically set as the value used for the model optimization with the source training images.

\hz{Our proposed dynamic learning rate adjustment at test time can be easily deployed to any network architecture and self-supervised objective function to improve the test-time adaptation process.}
Since the feature representation and prediction masks are naturally calculated and stored in the memory bank, the computational cost at test time mainly comes from the retrieval of support set for the calculation of reference prediction $\hat{p}_t$. We improve the efficiency by keeping a small retrieval size (e.g. 8). The effect of retrieval size is analyzed in the experiment and results show that a relatively small size can already achieve good results, increasing the size (e.g. increasing to 20) dose not show higher performance.

\algnewcommand{\LeftComment}[1]{\State  \(\triangleright\) #1 \hfill~}

\begin{algorithm}[!t]
\caption{\hz{Dynamic Learning Rate for Test-Time Adaptation (DLTTA) Learning Process}}
\label{alg:dltta}
\textbf{Input}: Test samples \hz{or batches $\{x^{}_{t,b}\}$ for $t$ from 1 to $T$ and $b$ from 1 to $B$}, model parameters learned from the source training data $\theta_{1}$, memory bank $\mathcal{M}$, learning rate for source train images $\alpha$, test-time batch size $B$\\
\textbf{Output}: The final prediction $\{\bar{y}_t\}_{t=1}^T$for test samples
\begin{algorithmic}[1] 
\State Initialize model $f=g \circ h$ with parameters $\theta_{1}$ 
\For{$t=1,2,\cdots,T$}
\For{$b=1, 2, \cdots, B$} 
\State $q_{t,b}=h(x_{t,b}), v_{t,b}=g(q_{t,b})$  

\Comment{Forward pass using $\theta_{t}$}
\State $\{(q_d, v_d)\}_{d=1}^{D} \sim \mathcal{M}$ \Comment{Retrieve support set}
\State $\hat{p}_{t,b}  = \frac{1}{D} \sum_{d=1}^{D} v_d$ \Comment{Eq.~(\ref{eq:p_t})}
\State $L_{\text{DIV},b}^{} = \frac{1}{2} (L_\text{KL}(\hat{p}_{t,b}||f_{\theta_t}(x_{t,b})) $ 

$\quad\quad\quad\quad\quad +L_\text{KL}(f_{\theta_{t}}(x_{t,b})||\hat{p}_{t,b}))$
 \Comment{Eq.~(\ref{eq:KL})}
\EndFor

\State $L_\text{DIV}^B  = \frac{1}{B} \sum_{b=1}^{B} L_{\text{DIV},b}$ \Comment{Eq.~(\ref{eq:batchKL})}

\State $\eta = \alpha \cdot L_\text{DIV}^B$ \Comment{Eq.~(\ref{eq:h_div})}
\State  $\theta_{t+1} \!\gets\! \theta_{t}\! -\! \eta ( \nabla L_{tt}f_{\theta_t}(x_t))$ \Comment{Dynamic learning}
\State $(q_t,v_t), \bar{y}_t = f_{\theta_{t+1}}(x_t)$ \Comment{Forward pass using $\theta_{t+1}$}
\State $\mathcal{M}.\text{add}((q_t,v_t))$ \Comment{Update the memory bank}
\EndFor
\State \textbf{return} $\{\bar{y}_t\}_{t=1}^t$ 
\end{algorithmic}
\end{algorithm}

\subsection{Learning Process and Implementation Details}
\subsubsection{Learning Process}
The overall procedure to perform our test-time adaptation is summarized as follows. 
The model to be adapted is initialized with parameters learned from the source training data. 
For a coming test sample, the model firstly performs a forward pass to obtain the pair of feature representation and prediction mask to retrieve semantically similar elements from the memory bank, and calculates the discrepancy and dynamic learning rate for the specific test sample. 
The model is then updated by using the derived learning rate with a test-time objective function to achieve desired adaptation. 
\hz{Note that only a partial model parameters are adapted following previous works with different test-time adaptation objectives~\cite{sun2020test,wang2020tent,he2021autoencoder}.}
After the one-step adaptation, the model performs another forward pass to obtain the prediction for the current test sample and update the new element pair for the memory bank. 
This model is iteratively updated with the previous process with each sequentially coming test sample.
Before the memory bank being constructed, the learning rate is kept as the initial value. The pseudo-code of DLTTA is shown in Algorithm~\ref{alg:dltta}.

\subsubsection{Implementation Details}
The choice of network architecture in our method is flexible. Without loss of generality, \hz{we followed \cite{he2021autoencoder} to employ a U-Net~\cite{he2019deep} for retinal OCT image segmentation, followed \cite{sagawa2021camelyon} to use a DenseNet-121~\cite{huang2017densely} pre-trained on ImageNet for histopathological image classification,
and employed a 3D U-Net~\cite{cciccek20163d_unet} for prostate MRI image segmentation}.
 \hz{The encoders of our 2D U-Net and 3D U-Net contain 4 Convolution-BatchNorm-ReLU blocks, which continuously down-sample the image resolution and double the feature channels dimensions. Then the features are processed by a bottleneck layer and fed into a decoder, which has 4 transpose convolution blocks to up-sample the intermediate feature maps. All 2D convolutions use kernel size 3×3 and 3D convolutions use kernel size 3×3×3. The encoder features are skip-connected to the decoder at each stage and the feature channel dimensions are [16, 32, 64, 128] for each block. For segmentation network, we consider the encoder and bottle layer as feature extractor and the decoder as the prediction head. The architecture of DenseNet-121 follows the implementation in Torchvision~\cite{paszke2019pytorch} library, which consists of 4 densely connected blocks followed by global pooling and fully connected classification layers.}
\hz{For each task, the network backbone of our method and other comparison approaches are the same to ensure fair comparison.}

For model training with the source images, model was trained from scratch for 100 epochs with Adam optimizer and learning rates were initialized as 1e-3 and 3e-4 for segmentation and classification tasks.
For test-time adaptation of our method and all the comparison baselines, we performed one-step adaption for each batch of test data with batch size 1 for segmentation and 200 for classification task. 
We stored key and value pairs for the latest $\frac{K}{B}$ adaptation steps. We empirically set $\frac{K}{B}$ as 20 for segmentation and 4 for classification. 
\hz{The retrieval size $D$ is set to 8 for segmentation task and 12 for classification task.}
Following~\cite{nado2020evaluating}, BN statistics were re-collected from test data for all the methods. 
The framework was implemented with Pytorch 1.7.0, and trained on one NVIDIA TitanXp GPU.

\renewcommand{\arraystretch}{1.3}
	\begin{table*}[]
		\centering
	    \caption{\hz{Comparison of different methods in retinal layer segmentation in Dice score on OCT test dataset of Cirrus images.}}
	    \resizebox{1.0\textwidth}{!}{%
	    \setlength\tabcolsep{4.0pt}
	    \scalebox{0.93}{
	    \begin{tabular}{l|cccccccc||c}
	       \hline
	      
	              Method
	             &RNFL &GCIP &INL &OPL  & ONL&IS  & OS &RPE& Average\\
	      \hline   
	      \hline
	  W/o Adaptation
	  & 64.61$\pm$0.08  & 68.88$\pm$0.02 & 62.34$\pm$0.06
	   & 56.64$\pm$0.03  & 83.03$\pm${0.02} & 85.51$\pm$0.03     & 88.27$\pm$0.03  & 82.85$\pm${0.03} & 74.01$\pm$0.05\\      
	 	  PTBN~\cite{nado2020evaluating}
	  &67.53$\pm$0.19 &80.29$\pm$0.19 &72.56$\pm$0.17
	  &61.93$\pm$0.09 & 85.42$\pm$0.14  &86.17$\pm$0.06 
	  &88.94$\pm$0.06 &84.03$\pm$0.04 &78.36$\pm$0.09\\
	  UDA-ST~\cite{gatys2016image}
	  &77.30$\pm$0.09 & 86.60$\pm$0.05 &78.20$\pm$0.05 
	   & 64.40$\pm$0.06 & 87.60$\pm$0.03 & 88.30$\pm$0.03
	  & 86.90$\pm$0.05 & 82.60$\pm$0.05 & 81.50$\pm$0.05\\
	  
	  \hz{ CycleGAN}~\cite{zhu2017cycgan}
	  &\hz{77.23$\pm$0.03} 
	  &\hz{86.49$\pm$0.10} 
	  &\hz{77.83$\pm$0.05 }
	  & \hz{63.88$\pm$0.10 }
	  & \hz{86.44$\pm$0.03 }
	  & \hz{88.27$\pm$0.01}
	  & \hz{86.64$\pm$0.05 }
	  & \hz{82.44$\pm$0.04 }
	  & \hz{81.15$\pm$0.06}\\
	\hz{UDAS}~\cite{Tsai2018udas}
	  &\hz{78.62$\pm$0.14 }
	  & \hz{87.04$\pm$0.13 }
	  &\hz{79.23$\pm$0.15 }
	   & \hz{65.09$\pm$0.08} 
	   & \hz{88.40$\pm$0.09 }
	   & \hz{88.58$\pm$0.07}
	  & \hz{86.42$\pm$0.11 }
	  & \hz{83.89$\pm$0.24 }
	  & \hz{82.16$\pm$0.12}\\              

	  \hline
	  TTT~\cite{sun2020test} 
	  &70.55$\pm$0.18 & 81.83$\pm$0.12 &74.14$\pm$0.16
	  &64.41$\pm$0.08 & 86.76$\pm$0.09 &86.85$\pm$0.12  
	  &89.15$\pm$0.08  & 84.27$\pm$0.08 &79.75$\pm$0.16
	  \\ 
        DLTTA (\textit{Ours})
	   & \textbf{71.09$\pm$0.14} & \textbf{82.97$\pm$0.12} &\textbf{75.30$\pm$0.18}
	    & \textbf{65.36$\pm$0.14} & \textbf{87.05$\pm$0.11}  & \textbf{88.31$\pm$0.06}
	    & \textbf{89.39$\pm$0.09} & \textbf{84.87$\pm$0.05}  &\textbf{80.54$\pm$0.11}
	  \\             
	  \hline
	  Tent~\cite{wang2020tent}  
	   &  76.95$\pm$0.13   &83.93$\pm$0.08  & 75.49$\pm$0.16  
	   &  66.76$\pm$0.13  & 88.10$\pm$0.12  &87.08$\pm$0.15    
	   & 89.28$\pm$0.13 &84.38$\pm$0.08  &81.50$\pm$0.15 \\
	 
    DLTTA (\textit{Ours})  
	   & \textbf{77.60$\pm$0.06}
	   & \textbf{84.27$\pm$0.04} 
	   & \textbf{75.78$\pm$0.07}  
	   & \textbf{69.08$\pm$0.05}  
	   & \textbf{90.38$\pm$0.06}
       & \textbf{88.42$\pm$0.09}
	   & \textbf{90.01$\pm$0.03}
	   & \textbf{85.61$\pm$0.10}
	   & \textbf{82.63$\pm$0.06}
	  \\
	  
	     \hline
	     ATTA~\cite{he2021autoencoder}
	  &73.20$\pm$0.13
	  &85.10$\pm$0.04
	  &79.10$\pm$0.03
	  &69.20$\pm$0.06 
	  &90.40$\pm$0.02
	  &88.40$\pm$0.03
	  &89.90$\pm$0.02
	  &84.90$\pm$0.03
	  &82.50$\pm$0.05
	  \\
    DLTTA (\textit{Ours})
	      &\textbf{73.86$\pm$0.08}
	     & \textbf{86.48$\pm$0.09}
	     &\textbf{79.86$\pm$0.04}
	     &\textbf{70.18$\pm$0.02}
	     &\textbf{91.57$\pm$0.07}
	     &\textbf{89.69$\pm$0.04}
	     &\textbf{90.12$\pm$0.05}
	     &\textbf{85.56$\pm$0.03}
	     &\textbf{83.41$\pm$0.05}
	  \\
	  \hline
	    
	    \end{tabular}}}
	    \begin{tablenotes}
		\item[*] \hz{RNFL: Retina nerve fiber layer, GCIP: Ganglion cell layer and inner plexiform layer, INL: Inner nuclear layer, OPL: Outer plexiform layer, ONL: Outer nuclear layer, IS: Inner photoreceptor segments, OS: Outer photoreceptor segments, RPE: Retinal pigment epithelium}
	    \end{tablenotes}
	    \label{tab:comparison-seg}
	\end{table*}
\section{Experiments}

\subsection{Datasets and Evaluation Metrics}
We first validate the effectiveness of our method on retinal layer segmentation with optical coherence tomography (OCT) datasets that present distribution shift caused by the different Heidelberg Spectralis scanner and Cirrus scanner~\cite{he2021autoencoder, he2019retinal}. Then we evaluate our method on a much larger dataset of histopathological images for tumor and normal tissue classification, that are obtained from different hospitals with varying protocols and patient populations~\cite{koh2021wilds, bandi2018detection, litjens2018camelyon}. 
\hz{We also validate our method on prostate segmentation with 3D MRI images collected from different medical centers~\cite{liu2020shape,nciisbi2013,litjens2014evaluation_prostate,lemaitre2015prostate}.}

\textbf{Retinal OCT Image Segmentation.}
We follow the literature~\cite{he2021autoencoder} to employ a public OCT dataset~\cite{he2019retinal} that is acquired from Heidelberg Spectralis scanners as the source training data, and a public dataset of OCT images obtained from Cirrus scanners~\cite{he2021autoencoder} as the shifted test dataset.
The source training dataset consists of 35 3D volumes, but due to the large physical distance between slices, following \cite{he2019retinal}, the 3D volumes were split into 588 2D slices for training, 147 slices for validation, and 980 slices for testing. 
The task is to segment eight different retinal layers with ground truth being provided for model training. 
The test dataset contains 48 slices from 6 subjects. 
As in \cite{he2021autoencoder}, the Cirrus 2D slices were re-sampled to have the same physical within-slice resolution as the Spectralis scans. All the slices from both datasets were pre-processed with retina flattening and cropped and resized to 128 $\times$ 1024. 
For evaluation, we follow previous works~\cite{he2021autoencoder} of using the same two OCT datasets to adopt the popular Dice score to evaluate the segmentation performance. 

\textbf{Histopathological Image Classification}.  
We use the large Camelyon17~\cite{litjens2018camelyon,koh2021wilds, bandi2018detection} dataset to evaluate our method, which contains more than 400,000 histopathological image patches of size 96$\times$96 from five different hospitals. 
In our experiment, 302,436 tissue patches collected from three hospitals are used as the source training data, 34,904 patches and 85,054 patches from two other hospitals are taken as two unseen test datasets respectively, i.e., unseen dataset A and unseen dataset B. The task is to predict whether a given region of tissue contains any tumor tissue. 
Following \cite{koh2021wilds}, we normalized each image patch to zero mean and unit variance with the data statistics of ImageNet.
Five most commonly used evaluation metrics of classification task, including accuracy, sensitivity, specificity, AUC, and F1 score are adopted to evaluate different methods. 

\hz{\textbf{Prostate MRI Image Segmentation.}
We further validate our approach on the prostate segmentation task across multi-center 3D MRI data. We use prostate T2-weighted MRI volumes of three public datasets NCI-ISBI13~\cite{nciisbi2013}, I2CVB~\cite{lemaitre2015prostate}, and PROMISE12~\cite{litjens2014evaluation_prostate}, which were collected from six clinical centers. We follow the previous work~\cite{liu2020shape} to partition the data into six datasets A to F, according to the clinical centers that the datasets were collected from. Dataset A is employed as the training data and the other datasets are used as different unseen test datasets. For data pre-processing, each MRI volume was normalized to zero mean and unit variance in intensity values and resized to 384*384 in the axial plane. Patches with size of 80*80*80 were cropped as the network inputs. 
The common evaluation metric Dice score is adopted to compare the performance of different methods.}

\subsection{Comparison with State-of-the-art Methods}
Our dynamic learning strategy can be deployed to different test-time adaptation methods to improve the effectiveness of model update. 
For each task, we deploy our dynamic learning rate strategy to three representative test-time adaptation methods, including \textbf{TTT}~\cite{sun2020test} which adapts the model with a proxy task of rotation prediction, \textbf{Tent}~\cite{wang2020tent} which adjusts the batch normalization layers by minimizing the entropy of model predictions on test data,
and \hz{\textbf{ATTA}~\cite{he2021autoencoder} which reduces domain shift with the autoencoders' reconstruction loss.}
We also compare our method with \textbf{PTBN}~\cite{nado2020evaluating} which re-estimates the batch normalization statistics from the test data.
To provide strong baselines, we include UDA methods in comparison that concurrently utilize source training data and the entire test dataset for adaptation. 
\hz{For retinal OCT image segmentation, we include three UDA methods that have been employed to adapt OCT images in \cite{he2021autoencoder}, that are the image translation methods \textbf{UDA-ST}~\cite{gatys2016image} and \textbf{CycleGAN}~\cite{zhu2017cycgan}, and the widely used output space adaptation method \textbf{UDAS}~\cite{Tsai2018udas}. 
For histopathological image classification, we compare with \textbf{UDA-SwAV}~\cite{sagawa2021camelyon}, which is the best reported UDA model for histopathological image classification using the datasets as ours. 
For prostate segmentation, we also employ \textbf{UDAS} for comparison. 
The results of ATTA, UDA-ST on OCT images and the results of UDA-SwAV on unseen histopathological dataset A are directly referenced from \cite{he2021autoencoder} and \cite{sagawa2021camelyon} respectively since the same datasets, network backbones, and data split are used in their methods and ours. 
The other results are obtained by re-implementing based on the released code with the network backbone being consistent for all the comparison methods.}
In addition, the \textbf{W/o Adaptation} model denotes directly applying the model optimized with training dataset to obtain the predictions of test samples.

\begin{table*}[]
    \renewcommand{\arraystretch}{1.3}
	\centering
	    \caption{Comparison of different methods in histopathological image classification on two different unseen test datasets.}
	    \resizebox{1.0\textwidth}{!}{%
	    \setlength\tabcolsep{3.0pt}
	    \scalebox{0.90}{
	    \begin{tabular}{l|ccccc||ccccc}
	       \hline
	        \multirow{2}{*}{Method} 
	        &\multicolumn{5}{c||}{Unseen Test Dataset A}  &\multicolumn{5}{c}{Unseen Test Dataset B}\\
	       \cline{2-11}
	        
	                    &Accuracy     &Sensitivity    &Specificity   &AUC    &F1
	                    &Accuracy     &Sensitivity    &Specificity   &AUC    &F1\\
	      	\hline
	      	\hline
	        {W/o Adaptation} &63.15$\pm${2.21} &64.65$\pm${1.73} &61.62$\pm${3.17} &80.72$\pm${2.24} &60.13$\pm${1.78}
	        &84.37$\pm${2.32} &72.30$\pm${2.15} &71.32$\pm${2.63} &81.98$\pm${1.33} &64.04$\pm${1.56}\\
	   {PTBN~\cite{nado2020evaluating}}
	         &86.17$\pm$0.44 &85.20$\pm$0.30 &85.08$\pm$0.93 &90.68$\pm$0.90
	         &86.58$\pm$0.15
	         &85.36$\pm$1.54 &85.79$\pm$1.01 &85.03$\pm$0.76 &89.93$\pm$0.65 &86.18$\pm$0.54
	            \\
	       {UDA-SwAV~\cite{sagawa2021camelyon}} 
	        &91.40$\pm$2.00 &92.39$\pm$1.67 & 89.54$\pm$1.84& 96.24$\pm$1.61
	         & 90.96$\pm$1.23
	         &88.64$\pm$1.25 & 90.29$\pm$1.72 & 88.91$\pm$1.87  & 95.24$\pm$2.31 &90.67$\pm$1.46 
	         \\

	   \hline
	      {TTT~\cite{sun2020test}}
	      
	        &87.74$\pm$0.60 &88.48$\pm$0.53 &86.90$\pm$0.68 &93.44$\pm$0.61
	         &87.79$\pm$0.14
	         &86.33$\pm$0.25 &87.61$\pm$0.62 &86.23$\pm$0.31 &92.31$\pm$0.29 &86.85$\pm$0.81
	         \\
	          {DLTTA (\textit{Ours})}  &{\textbf{89.03$\pm${0.95}}} &\textbf{{89.51}$\pm${0.64}} &\textbf{88.24$\pm${0.75}}
	          &\textbf{94.67$\pm${0.69}} &\textbf{{88.19}$\pm${0.42}}

	         &\textbf{{87.69}$\pm${0.25}} &\textbf{{88.52}$\pm${0.34}} 
	         &\textbf{87.74$\pm${0.39}} 
	         &\textbf{{93.65}$\pm${0.46}} &\textbf{{87.03}$\pm${0.26}}\\
	       \hline
	       {Tent~\cite{wang2020tent}}  
	         &90.17$\pm${0.50} &91.40$\pm${1.85} &\textbf{90.13$\pm${0.51}} &95.91$\pm${0.74} 
	         &91.00$\pm$0.29
	         &87.90$\pm$0.40 &89.50$\pm$0.27 &\textbf{88.86$\pm$0.76} &94.27$\pm$0.85 &89.75$\pm$0.38
	          \\
	      {DLTTA (\textit{Ours})}  &\textbf{91.49}$\pm$\textbf{0.07} &\textbf{92.68}$\pm$\textbf{0.12} &89.66$\pm${0.21} &\textbf{96.36}$\pm$\textbf{0.08} &\textbf{91.28}$\pm$\textbf{0.05}
	         &\textbf{88.93}$\pm$\textbf{0.19} &\textbf{90.59}$\pm$\textbf{0.14} 
	         &88.62$\pm${0.28} 
	         &\textbf{95.15}$\pm$\textbf{0.06} &\textbf{90.58}$\pm$\textbf{0.13}\\
	         
	  \hline   
	          {\hz{ATTA}~\cite{he2021autoencoder}} 
	        &  \hz{89.69$\pm$1.24}
	        &  \hz{90.89$\pm$1.99}
	        &  \hz{90.56$\pm$1.73}
	        &  \hz{94.23$\pm$1.22}
	        &  \hz{89.17$\pm$1.17}
	        &  \hz{87.54$\pm$1.23}
	        &  \hz{89.03$\pm$1.67}
	        &  \hz{87.42$\pm$1.69}
	        &  \hz{92.58$\pm$1.86}
	        &  \hz{88.52$\pm$1.74} 
	        \\
	         {\hz{DLTTA (\textit{Ours})}} 
	        & \hz{\textbf{91.30$\pm$1.51}} 
	        & \hz{\textbf{91.31$\pm$1.10}}
	        & \hz{\textbf{91.92$\pm$1.53}}
	        & \hz{\textbf{95.20$\pm$1.54}}
	        & \hz{\textbf{90.83$\pm$1.58}}
	        & \hz{\textbf{88.41$\pm$1.03}}
	        & \hz{\textbf{91.82$\pm$1.25}}
	        & \hz{\textbf{88.98$\pm$1.83}} 
	        & \hz{\textbf{94.28$\pm$1.61}} 
	        & \hz{\textbf{89.73$\pm$1.34}} 
	        \\

	   \hline
	    
	    \end{tabular}
	    }}
	    \label{tab:comparison-cls}
\end{table*}

\subsubsection{Results for Retinal Layer Segmentation}
\renewcommand{\arraystretch}{1.3}
	\begin{table*}[!h]
		\centering
	    \caption{\hz{Comparison of different methods in prostate segmentation on multi-center MRI datasets}}
	    \resizebox{0.75\textwidth}{!}{%
	    \setlength\tabcolsep{6.0pt}
	    \scalebox{0.93}{
	    \begin{tabular}{l|ccccc||c}
	       \hline
	       
	       \multirow{2}{*}{Method} 
	        &\multicolumn{5}{c||}{Dice Score in Unseen Test Datasets}  & \multirow{2}{*}{Average} \\
	       \cline{2-6}

	             & Dataset B & Dataset C & Dataset D  & Dataset E & Dataset F & \\
	      \hline   
	      \hline
	  W/o Adaptation
	  & 80.66$\pm$2.73
	  & 70.90$\pm$2.41
	  & 73.85$\pm$3.83
	   & 57.02$\pm$2.82
	   & 81.84$\pm$2.49
	   & 72.85$\pm$2.85   \\      
	
	PTBN~\cite{nado2020evaluating}
	  &80.82$\pm$2.29
	  &72.65$\pm$2.20
	  &73.95$\pm$3.21
	  &61.33$\pm$2.61
	  &82.07$\pm$3.02
	  &74.16$\pm$2.66\\
	 
	{UDAS}~\cite{Tsai2018udas}
	  &81.23$\pm$1.68
	  &75.19$\pm$1.92
	  &76.75$\pm$2.03
	  &65.83$\pm$1.42
	  &86.58$\pm$1.94
	  &77.11$\pm$1.79
	\\         

	  \hline
	  TTT~\cite{sun2020test} 
	  &81.09$\pm$2.25
	  &73.41$\pm$3.73
	  &74.29$\pm$2.85
	  &62.58$\pm$1.97
	  &82.49$\pm$2.56
	  &74.77$\pm$2.67
	 
	  \\ 
	  DLTTA (\textit{Ours})
	   &\textbf{83.14$\pm$2.34}
	  &\textbf{73.63$\pm$2.11}
	  &\textbf{76.41$\pm$2.14}
	  &\textbf{65.54$\pm$2.84}
	  &\textbf{83.42$\pm$2.77}
	  &\textbf{76.44$\pm$2.44}
	   
	  \\             
	  \hline
	  Tent~\cite{wang2020tent}  
	   &81.36$\pm$1.57
	  &76.10$\pm$1.48
	  &75.02$\pm$3.99
	  &62.23$\pm$1.85
	  &81.19$\pm$3.63
	  & 75.17$\pm$2.50
	   \\
	 
	  DLTTA (\textit{Ours})  
	   &\textbf{82.47$\pm$2.20}
	  &\textbf{77.84$\pm$2.81}
	  &\textbf{76.97$\pm$3.05}
	  &\textbf{66.32$\pm$2.92}
	  &\textbf{82.90$\pm$3.81}
	  &\textbf{77.30$\pm$2.95}
	  \\
	  
	     \hline
	     ATTA~\cite{he2021autoencoder}
	  &81.19$\pm$3.60
	  &75.94$\pm$3.02
	  &74.49$\pm$3.04
	  &68.06$\pm$2.28
	  &85.09$\pm$2.17
	  &76.95$\pm$2.82
	  \\
	  
	  DLTTA (\textit{Ours})
	      &\textbf{82.13$\pm$2.47}
	  &\textbf{76.50$\pm$2.45}
	  &\textbf{75.33$\pm$3.57}
	  &\textbf{70.21$\pm$2.02}
	  &\textbf{86.25$\pm$2.09}
	  &\textbf{78.08$\pm$2.52}
	    
	  \\
	  \hline
	    
	    \end{tabular}
	    }}
	    \label{tab:comparison-prostate}
	\end{table*}
\begin{figure*}[]
	\centering
	\vspace{0mm}
	\includegraphics[scale=0.43]{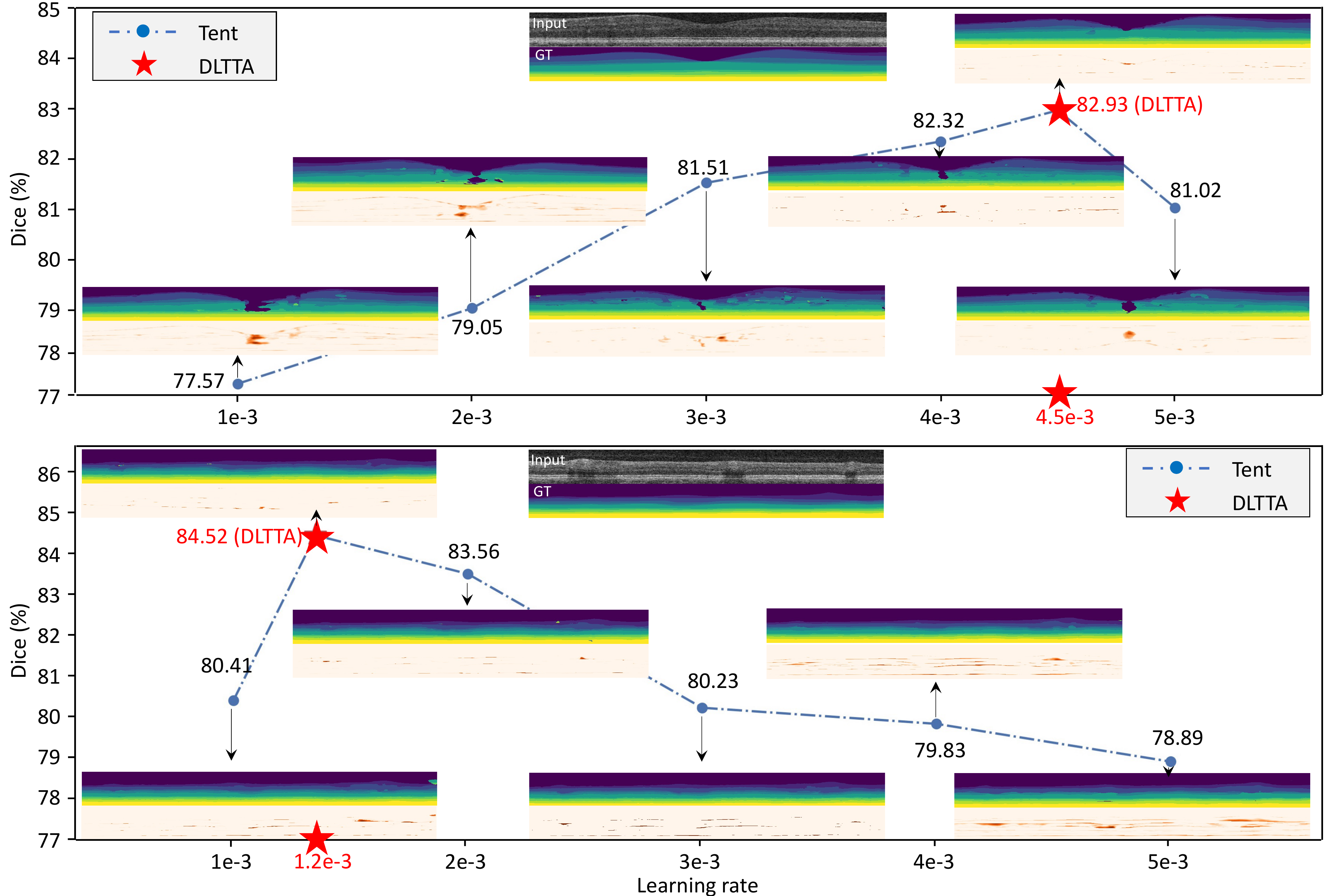}
	\vspace{-2mm}
	\caption{Dice score of the segmentation prediction produced by Tent with different static learning rates and our DLTTA, compared with the ground truth masks (GT). Each subplot aside the Dice value consists of the predicted segmentation masks (upper half, different colors represent different retinal layers) and the error map (lower half, the absolute difference between the prediction and ground truth).}	
	\label{fig:visual}
\end{figure*}

\textcolor{black}{
Table \ref{tab:comparison-seg} presents the comparison results for retinal layer segmentation from OCT test dataset of Cirrus images.
We can see that all test-time adaptation and UDA methods improve over the ``W/o Adaptation" baseline, showing the benefits of model adaptation towards the data distribution of test samples. 
Our DLTTA equipped with the dynamic learning rate strategy consistently improves different test-time adaptation methods TTT, Tent and ATTA for the segmentation of all retinal layers.
These results demonstrate the effectiveness of our dynamic learning rate on improving the test-time model update to overcome the varying distribution shift of test data, and \hz{validate that our dynamic learning rate strategy can be applied to different test-time adaptation methods to improve adaptation performance.} 
With deploying to ATTA, our method achieves the highest mean Dice values 83.41\%, which outperforms all UDA  methods \textcolor{black}{utilizing the source training data during adaptation process. 
Plain Tent obtains comparable performance to UDA-ST, but further adding our dynamic learning rate strategy improves the mean Dice from 81.50\% to 82.63\%, outperforming UDA-ST.} \hz{UDAS outperforms plain Tent but performs slightly worse than ATTA. When ATTA is equipped with the dynamic learning rate strategy, it further outperforms UDAS.}
\textcolor{black}{This shows the great potential of dynamic test-time adaptation and indicates that the source training data are not necessarily needed to achieve good adaptation performance. 
The possible reason could be that the dynamic test-time adaptation updates the model directly based on the distributional information of each specific test sample, instead of trying to find a domain-invariant space for the entire training and test data which could be difficult when the training and test images differ significantly in data distributions.}
}

\subsubsection{Results for Histopathological Image Classification}
Table \ref{tab:comparison-cls} shows the comparison results on the large histopathological image dataset for tumor and normal tissue classification. Results are reported on two different test datasets. 
We can see that even though the model is trained on images from three hospitals covering multiple data distributions, the model still performs poorly on the two unseen datasets with distribution shift, obtaining merely 63.15\% accuracy on unseen dataset A. 
This shows the necessity of model adaptation towards the distribution of test data. 
Similar to the retinal layer segmentation, our dynamic learning rate strategy also consistently improves TTT, Tent and \hz{ATTA} on the histopathological image classification task, showing the general applicability of our method in terms of test-time adaptation objectives, analysis tasks, and network backbones. 
By deploying to Tent, our method obtains 91.49\% and 88.93\% accuracy for the two test datasets respectively, which are comparable to the strong baseline of UDA-SwAV method utilizing both training and test data. 
The improvements are benefited from our adaptive learning rate adjustment to explicitly modulate the online adaptation process according to the estimated discrepancy for each test sample.

\hz{\subsubsection{Results for Prostate Segmentation}
Table \ref{tab:comparison-prostate} presents the prostate segmentation results with Dice score for MRI data from the five unseen test datasets B to F. We can see that the model learned from the training dataset obtains varying performance across the five test datasets due to distribution shift. Similar to the observations in retinal layer segmentation} \hz{and histopathological image classification, all test-time adaptation methods improve over the “W/o Adaptation” baseline.} \hz{With the proposed dynamic learning rate strategy, our DLTTA consistently improves TTT, Tent and ATTA methods for all the test datasets. By deploying dynamic learning rate to ATTA, our method achieves 78.08\% average Dice score, outperforming the unsupervised domain adaptation method UDAS which requires concurrent access to both training and test datasets. The results validate that our method is generally applicable to improve test-time adaptation with 3D models.
}

\subsection{Ablation Analysis of Our Method}
\textcolor{black}{We conduct ablation studies to investigate several important questions regarding our \hz{dynamic learning rate strategy}: 1) effectiveness of our learning rate adjustment, 2) how the adaptation stability is influenced by our method, 3) effect of initial learning rate, 4) influence of retrieval size $D$,} \hz{ 5) influence of image orders, 6) effect of similarity metrics.}

\subsubsection{Effectiveness of Our Learning Rate Adjustment}
\textcolor{black}{We first show that different test images require different learning rates to achieve better adaptation results. 
As shown in Fig.~\ref{fig:visual}, for the first test case, Tent with the static learning rate 4e-3 obtains better retinal layer segmentation results than the other learning rates, while for the second test case, Tent with the learning rate 1e-3 obtains higher Dice score.
This validates our motivation that a fixed learning rate can hardly suit all the test samples.
Our dynamic learning strategy adaptively sets the learning rate as 4.5e-3 for the model update on the first test case, and 1.2e-3 for the second one, obtaining higher segmentation results for both test samples.
This shows the effectiveness of our discrepancy estimation-based learning rate adjustment.}

\subsubsection{Adaptation Stability}
\begin{figure}[]
		\centering
	\vspace{-2mm}	\includegraphics[scale=0.35]{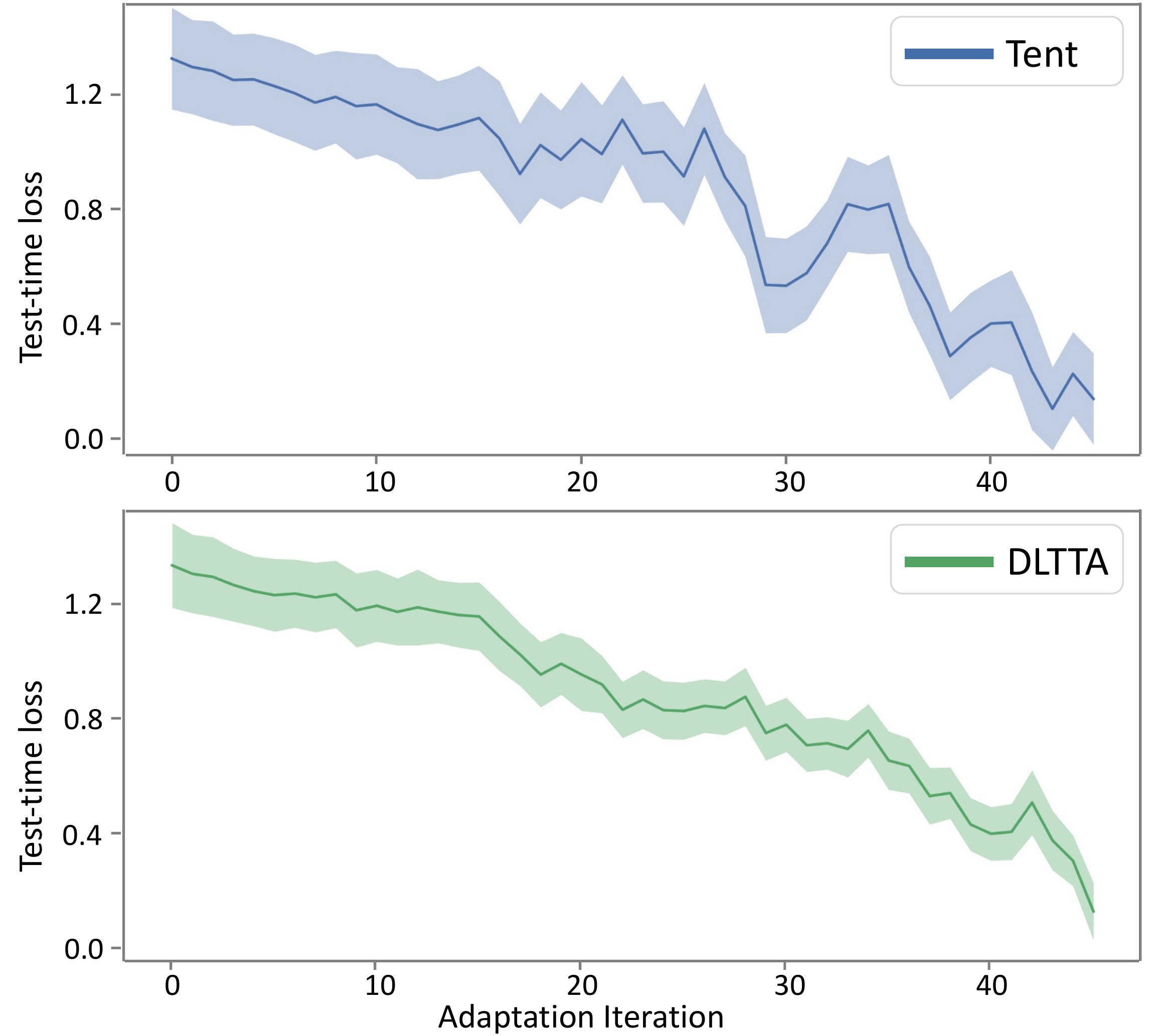}
	\vspace{-2mm}
		\caption{Test-time loss of Tent (upper) and our DLTTA (bottom) on retinal layer segmentation task. 
		Each plot shows the mean and standard deviation loss over all pixels of a test image, as a function of the test-time adaptation iterations.
		Our dynamic learning can stabilize the test-time adaptation process with smoother loss curve.}
		\vspace{-2mm}
		\label{fig:losscurve}
\end{figure}

\textcolor{black}{We compare the adaptation stability of Tent and our dynamic learning on the retinal layer segmentation task. Fig.~\ref{fig:losscurve} plots the test-time loss change with the adaptation iterations. We can see that the test-time loss curve of Tent fluctuates largely between the 25 to 40 training iterations. This may because that the fixed learning rate used in Tent cannot suit the varying adaptation demand of different test samples, leading to under- or over-adaptation for some test data. 
With our adaptive learning rate adjustment, the adaptation is dynamically tailored for each test sample, hence producing more stable adaptation with smoother loss curve.}

\subsubsection{Effect of Initial Learning Rate}
\textcolor{black}{We investigate the effect of different initial learning rates on test-time adaptation methods TTT, Tent, ATTA, and our dynamic adaptation.
As shown in Table \ref{tab:intial-lr}, with different initial learning rates from the set \{1e-3, 2e-3, 3e-3, 4e-3, 5e-3\}, our dynamic learning strategy consistently improves TTT, Tent and ATTA on Dice score in the retinal layer segmentation from OCT images. 
This indicates the general benefits of our dynamic learning rate adjustment regarding different initial learning rates. 
When deploying to Tent, our dynamic adaptation obtains over 82\% Dice score for all different initial learning rates, which is more stable than Tent. 
This shows that the performance of our dynamic adaptation method is less sensitive to the choice of initial learning rate compared to previous TTA methods which adopt a fixed learning rate for all the test-time adaptation steps.}

\subsubsection{The Influence of Retrieval Size}
\begin{figure}[]
		\centering
		\includegraphics[scale=0.27]{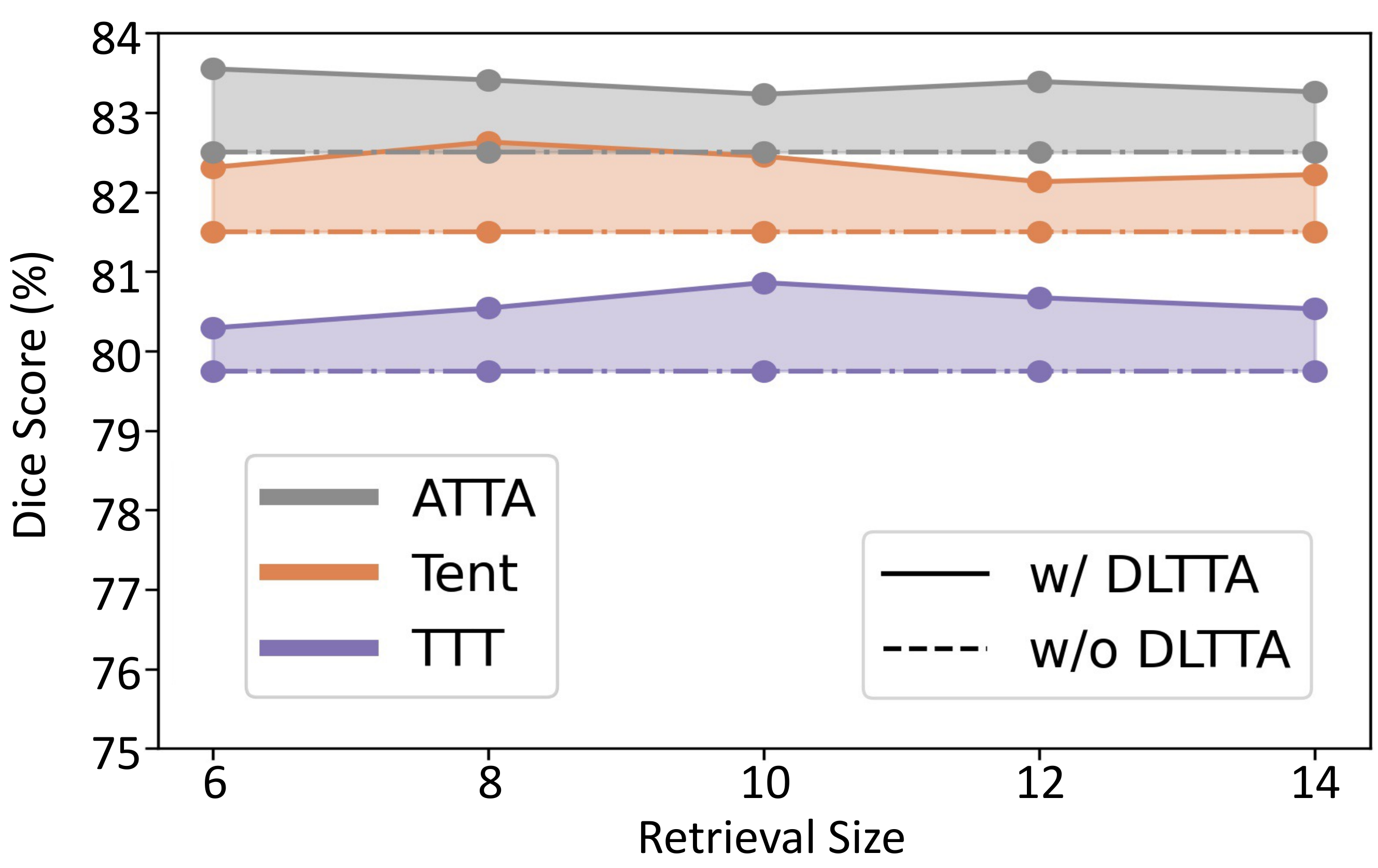}
		\vspace{-2mm}
		\caption{\hz{The change of Dice score on retinal layer segmentation with different retrieval size of elements from the memory bank.}}
		\vspace{-2mm}
		\label{fig:retrieve}
\end{figure}

\textcolor{black}{We study how the retrieval size ${D}$ for discrepancy estimation affects the performance of our method. 
Intuitively, less elements retrieved from the memory bank (i.e., smaller $D$) might be difficult to find sufficient semantically similar samples to provide prediction reference, while too more elements (i.e., larger $D$) could include some less relevant samples. 
Fig.~\ref{fig:retrieve} shows the change of Dice score on retinal layer segmentation when varying retrieval sizes $D\in \{6,8,10,12,14\}$. 
With different retrieval size, our method always obtains better performance than TTT, Tent and ATTA with the static learning strategy. 
The models with middle-level dictionary size ($D=8$) perform generally better for all the three TTA objectives than the model with smaller or larger retrieval size.}

\begin{table}[]
	\centering
    \caption{\hz{Comparison of test-time adaptation methods with different initial learning rates for retinal OCT image segmentation.}}
  
    \scalebox{1.0}{
    \begin{tabular}{lccccc}
        \toprule
         Method      &1e-3     &2e-3    &3e-3   &4e-3  &5e-3    \\
        \midrule
        
     TTT~\cite{sun2020test} &79.75  &80.45 &80.24 & 79.43&79.86     \\
     DLTTA (\textit{Ours})  &\textbf{80.54} &\textbf{81.47} &\textbf{81.09} & \textbf{80.41}&\textbf{80.67}     \\
    \hline
    Tent~\cite{wang2020tent}  &81.50 &80.90 &81.61 & 81.68&81.25     \\
     
     DLTTA (\textit{Ours})  &\textbf{82.63} & \textbf{82.36} & \textbf{82.48} &\textbf{82.46} &  \textbf{82.27}      \\
     
     \hline
     \hz{ATTA}~\cite{he2021autoencoder} &82.50 &82.77 &83.21 &82.42 &82.21\\
     \hz{DLTTA} (\textit{Ours}) &\textbf{83.41} &\textbf{83.45} &\textbf{83.59} &\textbf{83.35} &\textbf{83.30}\\
 
     \bottomrule
     
    \end{tabular}
    }
    \label{tab:intial-lr}
\end{table}

\subsubsection{Influence of Image Orders}
\hz{As the test images are considered to arrive as a sequence, we study whether the image orders affect the adaptation performance. We run different test-time \hz{adaptation methods on retinal layer segmentation with five different random orders of test images. The test dataset} \hz{is shuffled before the start of the online adaptation and the} \hz{image order is shared across all methods. 
The results presented in Table \ref{tab:order} show that the performance of our method as well as other test-time adaptation methods are not sensitive to the image orders. For our method, the largest variation is merely 0.19 in Dice.}
}
 
\subsubsection{Effect of Similarity Metrics}
\hz{
Since images in the memory bank are retrieved based on relative similarity order, choices} \hz{of specific similarity metrics can be flexible as long as the relative order is maintained. 
We have compared the results of our method between using L2 distance and consine similarity metrics on retinal layer segmentation and the dice performance difference is less than 0.1\%. 
This shows that our method is not sensitive to the choice of similarity metric. 
We adopt L2 distance in this work, but other metrics such as consine similarity metric can also be employed.} 

\renewcommand{\arraystretch}{1.3}
\begin{table}[]
\centering
\caption{\hz{Average Dice score of different methods on retinal layer segmentation with five random image orders.}}
\resizebox{0.45\textwidth}{!}{%
\setlength\tabcolsep{4.0pt}
\scalebox{0.93}{
\begin{tabular}{lccccc}
        \toprule
         Method      &Order A   &Order B   &Order C  &Order D   &Order E   \\
        \midrule
        
    TTT~\cite{sun2020test}  &79.75 &79.56 &79.58 &79.82 &79.77      \\
     DLTTA (\textit{Ours})  &\textbf{80.54} &\textbf{80.43} &\textbf{80.62} &\textbf{80.48}  &\textbf{80.57}    \\    
    \hline
     Tent~\cite{wang2020tent}  &81.50 &81.57 &81.45  &81.53 &81.49    \\
     DLTTA (\textit{Ours})  &\textbf{82.63} &\textbf{82.55} &\textbf{82.67} &\textbf{82.58} &\textbf{82.51}     \\
    \hline
    ATTA~\cite{he2021autoencoder}  &82.50  &82.52 &82.34 &82.42 &82.44   \\
     
     DLTTA (\textit{Ours})  &\textbf{83.41} & \textbf{83.42} & \textbf{83.35} & \textbf{83.29}   & \textbf{83.36}   \\
 
     \bottomrule
     
    \end{tabular}
    }}
    \label{tab:order}
\end{table}

\section{Discussions}

\textcolor{black}{Deep neural networks are notoriously difficult to generalize to unseen domains due to data distribution shift caused by varying image acquisition conditions. 
This paper tackles the challenging problem of test-time adaptation, which aims to generalize the deep models to unknown data distributions by learning from the inference sample provided at test-time.}
\hz{Previous methods on test-time adaptation utilize a fixed learning rate for all test samples. This solution is sub-optimal since the amount that the weights should be updated could be different for the sequentially arriving test images with varying degree of distribution shift. }
\hz{To address this problem, in this work, we propose a dynamic learning rate strategy for test-time adaptation, aiming to dynamically adjust the step size of model update according to the estimated prediction discrepancy.}
\textcolor{black}{Our proposed dynamic learning rate adjustment is general and can be easily applied to different test-time adaptation methods to improve performance. 
The general applicability and effectiveness of our method has been validated on three popular test-time objective functions, both segmentation and classification tasks, three imaging modalities, 2D and 3D models, and different network architectures. }

\hz{For model learning with the gradient descent optimization algorithm, learning rate determines the step size at each iteration thus controls how much to update the weights. Learning rate is an important hyperparameter and requires careful selection and scheduling even in the standard supervised learning~\cite{goyal2017step_decay,loshchilov2016anneal,maclaurin2015hyper_gradient,baydin2018online_hypergradient}. We consider the dynamic adjustment of learning rate becomes more significant during test-time adaptation, where a model needs to take suitable adaptation pace for the encountered new environment. 
In gradient descent optimization, the weights update is determined by the learning rate and the gradient of loss function. The main role of gradient of loss function is to indicate how the loss value would change given the changes of the model weights, thus determines the update direction. Although the amount of model updates can be also affected by the magnitude of gradient to some extent, it is mainly controlled by the learning rate. Therefore, loss function and learning rate play complementary role in the model learning process and both the two factors require careful design to achieve better model adaptation at test time. }

\hz{Our method follows the continual online adaptation setting, where the model adaptation on test image $x_t$ is based on the model parameters updated on the previous test sample $x_{t-1}$. Compared to image-specific adaptation~\cite{wang2018interactive}, where the model is re-initialized each time with the source model parameters, continual adaptation has shown better performance in previous work~\cite{sun2020test}. This may benefit from that continual adaptation allows the model to exploit the information provided by both current image and previous images. We also implement the image-specific variants of TTA methods. Each test image is updated with 10 iterations to allow sufficient adaptation. 
In Table \ref{tab:specific}, we obtain similar observation to \cite{sun2020test} that the continual adaptation outperforms the image-specific version and takes 10 times less computation. 
Our method can benefit more for the continual adaptation by dynamically modulating the online model update.}

\hz{Existing TTA methods either take multiple adaptation steps for each test image, such as ATTA, or one gradient update, such as TTT and Tent. For the adaptation of each test image with our method, we only perform one gradient update because benefited from the proposed dynamic learning rate adjustment to explicitly module the online adaptation process, one single step can already achieve effective adaptation.
We also try to update the model for multiple steps on each test image. 
Table \ref{tab:updation} presents the results of our methods with 1, 4, and 8 gradient updates. We can see that no consistent improvements can be obtained with multiple updates but taking more computation cost. These results demonstrate that our method can achieve effective and efficient test-time adaptation with just one gradient update, requiring less computation cost but achieving improved performance than previous methods.}

\begin{table}[]
	\centering
    \caption{\hz{Average Dice score of retinal OCT image segmentation with different test-time adaptation settings.
}}
  \resizebox{0.49\textwidth}{!}{%
\setlength\tabcolsep{3.5pt}
\scalebox{1.0}{
    \begin{tabular}{l|cc|cc|cc}
        \hline
         Adaptation Setting  &TTT &DLTTA     &Tent &DLTTA  &ATTA &DLTTA       \\
        \hline
        Image-specific  
     &76.40 &\textbf{76.83}
     &81.23 &\textbf{81.79} 
     &81.85 &\textbf{82.56}      \\
     
     Image-continual  
     &79.75 &\textbf{80.54} 
     &81.50 &\textbf{82.63} 
     &82.50 &\textbf{83.41}  \\
     
    \hline
     
    \end{tabular}
    }}
    \label{tab:specific}
\end{table}

\hz{One limitation of our method is that before the memory bank being constructed, it is difficult to estimate the discrepancy, thus the dynamic learning rate adjustment is not added to the test-time adaptation of the first K images. Without using dynamic learning rate for adaptation, the performance of the first K images in our method is the same as that in previous TTA methods depending on the objectives functions that the dynamic learning rate is deployed to. It would be interesting future work to explore how to obtain better discrepancy estimation for dynamic learning rate adjustment of the first several test images and to achieve faster and more effective adaptation at the beginning. A possible solution could be storing some training data feature statistics in the memory bank and adjust the learning rate by comparing training and testing feature statistics and predictions.}

\begin{table}[]
	\centering
    \caption{\hz{Average Dice score of our method on retinal layer segmentation with different number of update steps.}}
    \resizebox{0.33\textwidth}{!}{%
\setlength\tabcolsep{4.0pt}
  
    \scalebox{1.0}{
    \begin{tabular}{lccc}
        \hline
         Method      &1   &4   &8  \\
        \hline
        
    DLTTA based on TTT  &80.54 &80.69 &80.67 \\
        
     DLTTA based on Tent  &82.63 &82.41 &82.35 \\
     
     DLTTA based on ATTA  &83.41 &83.45 &83.23 \\
 
     \hline
     
    \end{tabular}
    }}
    \label{tab:updation}
\end{table}

\textcolor{black}{In general, our proposed method of dynamic learning rate strategy can be applied to many test-time learning algorithms. In this work, we mainly consider the rotation prediction, entropy minimization and autoencoder reconstruction as the objective to drive the model adaptation, while our idea to adopt dynamic step size is independent from the objective choices. \hz{For future work, we are interested in investigating the effectiveness of our method in more challenging adaptation scenarios, such as when test images come from multiple continual changing domains.} }

\section{Conclusion}
\textcolor{black}{We present the first method for dynamic learning rate adjustment of test-time adaptation to effectively adapt the model towards the varying distribution shift among test data. 
We propose a memory bank-based discrepancy measurement by considering both the progressive model change at test time and the variations in test data distributions, and further achieve dynamic learning rate adjustment based on the estimated discrepancy.
Our method is effective and generic to test-time adaptation objectives, requiring no change on the network designs, thus can be readily applied to improve different test-time adaptation methods. }

 \normalem
 \textcolor{black}{\bibliography{IEEEexample.bib}}

\begin{thebibliography}{10}
\providecommand{\url}[1]{#1}
\csname url@samestyle\endcsname
\providecommand{\newblock}{\relax}
\providecommand{\bibinfo}[2]{#2}
\providecommand{\BIBentrySTDinterwordspacing}{\spaceskip=0pt\relax}
\providecommand{\BIBentryALTinterwordstretchfactor}{4}
\providecommand{\BIBentryALTinterwordspacing}{\spaceskip=\fontdimen2\font plus
\BIBentryALTinterwordstretchfactor\fontdimen3\font minus
  \fontdimen4\font\relax}
\providecommand{\BIBforeignlanguage}[2]{{%
\expandafter\ifx\csname l@#1\endcsname\relax
\typeout{** WARNING: IEEEtran.bst: No hyphenation pattern has been}%
\typeout{** loaded for the language `#1'. Using the pattern for}%
\typeout{** the default language instead.}%
\else
\language=\csname l@#1\endcsname
\fi
#2}}
\providecommand{\BIBdecl}{\relax}
\BIBdecl

\bibitem{zhao2020review}
S.~Zhao, X.~Yue, S.~Zhang, B.~Li, H.~Zhao, B.~Wu, R.~Krishna, J.~E. Gonzalez,
  A.~L. Sangiovanni-Vincentelli, S.~A. Seshia \emph{et~al.}, ``A review of
  single-source deep unsupervised visual domain adaptation,'' \emph{IEEE
  Transactions on Neural Networks and Learning Systems}, 2020.

\bibitem{kamnitsas2017unsupervised}
K.~Kamnitsas, C.~Baumgartner, C.~Ledig, V.~Newcombe, J.~Simpson, A.~Kane,
  D.~Menon, A.~Nori, A.~Criminisi, D.~Rueckert \emph{et~al.}, ``Unsupervised
  domain adaptation in brain lesion segmentation with adversarial networks,''
  in \emph{International Conference on Medical Image Computing and Computer
  Assisted Intervention}.\hskip 1em plus 0.5em minus 0.4em\relax Springer,
  2017, pp. 597--609.

\bibitem{zhang2018task}
Y.~Zhang, S.~Miao, T.~Mansi, and R.~Liao, ``Task driven generative modeling for
  unsupervised domain adaptation: Application to x-ray image segmentation,'' in
  \emph{International Conference on Medical Image Computing and Computer
  Assisted Intervention}, 2018, pp. 599--607.

\bibitem{chen2020unsupervised}
C.~Chen, Q.~Dou, H.~Chen, J.~Qin, and P.~A. Heng, ``Unsupervised bidirectional
  cross-modality adaptation via deeply synergistic image and feature alignment
  for medical image segmentation,'' \emph{IEEE transactions on medical
  imaging}, vol.~39, no.~7, pp. 2494--2505, 2020.

\bibitem{ghafoorian2017transfer}
M.~Ghafoorian, A.~Mehrtash, T.~Kapur, N.~Karssemeijer, E.~Marchiori,
  M.~Pesteie, C.~R. Guttmann, F.-E. de~Leeuw, C.~M. Tempany \emph{et~al.},
  ``Transfer learning for domain adaptation in mri: Application in brain lesion
  segmentation,'' in \emph{International Conference on Medical Image Computing
  and Computer Assisted Intervention}, 2017, pp. 516--524.

\bibitem{gibson2018inter}
E.~Gibson, Y.~Hu, N.~Ghavami, H.~U. Ahmed, C.~Moore, M.~Emberton \emph{et~al.},
  ``Inter-site variability in prostate segmentation accuracy using deep
  learning,'' in \emph{International Conference on Medical Image Computing and
  Computer Assisted Intervention}.\hskip 1em plus 0.5em minus 0.4em\relax
  Springer, 2018, pp. 506--514.

\bibitem{huo2018synseg}
Y.~Huo, Z.~Xu, H.~Moon, S.~Bao \emph{et~al.}, ``Synseg-net: Synthetic
  segmentation without target modality ground truth,'' \emph{IEEE transactions
  on medical imaging}, vol.~38, no.~4, pp. 1016--1025, 2018.

\bibitem{DouOCCH18}
Q.~Dou, C.~Ouyang, C.~Chen, H.~Chen, and P.~Heng, ``Unsupervised cross-modality
  domain adaptation of convnets for biomedical image segmentations with
  adversarial loss,'' in \emph{International Joint Conferences on Artificial
  Intelligence}, 2018, pp. 691--697.

\bibitem{zhang2018translating}
Z.~Zhang, L.~Yang, and Y.~Zheng, ``Translating and segmenting multimodal
  medical volumes with cycle-and shape-consistency generative adversarial
  network,'' in \emph{CVPR}, 2018, pp. 9242--9251.

\bibitem{varsavsky2020test}
T.~Varsavsky, M.~Orbes-Arteaga, C.~H. Sudre, M.~S. Graham, P.~Nachev, and M.~J.
  Cardoso, ``Test-time unsupervised domain adaptation,'' in \emph{International
  Conference on Medical Image Computing and Computer Assisted
  Intervention}.\hskip 1em plus 0.5em minus 0.4em\relax Springer, 2020, pp.
  428--436.

\bibitem{zhang2018multi}
L.~Zhang, M.~Perea{\~n}ez, S.~K. Piechnik, S.~Neubauer, S.~E. Petersen, and
  A.~F. Frangi, ``Multi-input and dataset-invariant adversarial learning (mdal)
  for left and right-ventricular coverage estimation in cardiac mri,'' in
  \emph{International Conference on Medical Image Computing and Computer
  Assisted Intervention}.\hskip 1em plus 0.5em minus 0.4em\relax Springer,
  2018, pp. 481--489.

\bibitem{zhao2018supervised}
H.~Zhao, H.~Li, S.~Maurer-Stroh, Y.~Guo, Q.~Deng \emph{et~al.}, ``Supervised
  segmentation of un-annotated retinal fundus images by synthesis,'' \emph{IEEE
  transactions on medical imaging}, vol.~38, no.~1, pp. 46--56, 2018.

\bibitem{bateson2020source}
M.~Bateson, H.~Kervadec \emph{et~al.}, ``Source-relaxed domain adaptation for
  image segmentation,'' in \emph{International Conference on Medical Image
  Computing and Computer Assisted Intervention}.\hskip 1em plus 0.5em minus
  0.4em\relax Springer, 2020, pp. 490--499.

\bibitem{chen2021source}
C.~Chen, Q.~Liu, Y.~Jin, Q.~Dou, and P.-A. Heng, ``Source-free domain adaptive
  fundus image segmentation with denoised pseudo-labeling,'' in
  \emph{International Conference on Medical Image Computing and Computer
  Assisted Intervention}.\hskip 1em plus 0.5em minus 0.4em\relax Springer,
  2021, pp. 225--235.

\bibitem{tzeng2017adversarial}
E.~Tzeng, J.~Hoffman, K.~Saenko, and T.~Darrell, ``Adversarial discriminative
  domain adaptation,'' in \emph{CVPR}, 2017, pp. 7167--7176.

\bibitem{sun2020test}
Y.~Sun, X.~Wang, Z.~Liu, J.~Miller, A.~Efros, and M.~Hardt, ``Test-time
  training with self-supervision for generalization under distribution
  shifts,'' in \emph{ICML}.\hskip 1em plus 0.5em minus 0.4em\relax PMLR, 2020,
  pp. 9229--9248.

\bibitem{wang2020tent}
D.~Wang, E.~Shelhamer, S.~Liu, B.~A. Olshausen, and T.~Darrell, ``Tent: Fully
  test-time adaptation by entropy minimization,'' in \emph{ICLR}, 2021.

\bibitem{nado2020evaluating}
Z.~Nado, S.~Padhy, D.~Sculley, A.~D'Amour, B.~Lakshminarayanan, and J.~Snoek,
  ``Evaluating prediction-time batch normalization for robustness under
  covariate shift,'' \emph{arXiv preprint arXiv:2006.10963}, 2020.

\bibitem{he2020sdan}
Y.~He, A.~Carass, L.~Zuo, B.~E. Dewey \emph{et~al.}, ``Self domain adapted
  network,'' in \emph{International Conference on Medical Image Computing and
  Computer Assisted Intervention}, 2020, pp. 437--446.

\bibitem{karani2021test}
N.~Karani, E.~Erdil, K.~Chaitanya, and E.~Konukoglu, ``Test-time adaptable
  neural networks for robust medical image segmentation,'' \emph{Medical Image
  Analysis}, vol.~68, p. 101907, 2021.

\bibitem{he2021autoencoder}
Y.~He, A.~Carass, L.~Zuo, B.~E. Dewey, and J.~L. Prince, ``Autoencoder based
  self-supervised test-time adaptation for medical image analysis,''
  \emph{Medical Image Analysis}, p. 102136, 2021.

\bibitem{liu2021ttt++}
Y.~Liu, P.~Kothari, B.~van Delft, B.~Bellot-Gurlet, T.~Mordan \emph{et~al.},
  ``Ttt++: When does self-supervised test-time training fail or thrive?''
  \emph{Advances in Neural Information Processing Systems}, vol.~34, 2021.

\bibitem{iwasawa2021prototyp}
Y.~Iwasawa and Y.~Matsuo, ``Test-time classifier adjustment module for
  model-agnostic domain generalization,'' \emph{Neurips}, vol.~34, 2021.

\bibitem{pandey2021generalization}
P.~Pandey, M.~Raman, S.~Varambally, and P.~AP, ``Generalization on unseen
  domains via inference-time label-preserving target projections,'' in
  \emph{CVPR}, 2021, pp. 12\,924--12\,933.

\bibitem{ioffe2015batch}
S.~Ioffe and C.~Szegedy, ``Batch normalization: Accelerating deep network
  training by reducing internal covariate shift,'' in \emph{International
  conference on machine learning}.\hskip 1em plus 0.5em minus 0.4em\relax PMLR,
  2015, pp. 448--456.

\bibitem{chen2020simclr}
T.~Chen, S.~Kornblith, M.~Norouzi, and G.~Hinton, ``A simple framework for
  contrastive learning of visual representations,'' in \emph{International
  conference on machine learning}.\hskip 1em plus 0.5em minus 0.4em\relax PMLR,
  2020, pp. 1597--1607.

\bibitem{varsavsky2020testUDA}
T.~Varsavsky, M.~Orbes-Arteaga, C.~H. Sudre \emph{et~al.}, ``Test-time
  unsupervised domain adaptation,'' in \emph{International Conference on
  Medical Image Computing and Computer Assisted Intervention}.\hskip 1em plus
  0.5em minus 0.4em\relax Springer, 2020, pp. 428--436.

\bibitem{liu2021adapting}
X.~Liu, F.~Xing, C.~Yang, G.~El~Fakhri, and J.~Woo, ``Adapting off-the-shelf
  source segmenter for target medical image segmentation,'' in
  \emph{International Conference on Medical Image Computing and Computer
  Assisted Intervention}.\hskip 1em plus 0.5em minus 0.4em\relax Springer,
  2021, pp. 549--559.

\bibitem{zhu2021regis}
W.~Zhu, Y.~Huang, D.~Xu, Z.~Qian, W.~Fan, and X.~Xie, ``Test-time training for
  deformable multi-scale image registration,'' \emph{ICRA}, 2021.

\bibitem{liu2021generative}
X.~Liu, F.~Xing, M.~Stone, J.~Zhuo, T.~Reese, J.~L. Prince, G.~El~Fakhri, and
  J.~Woo, ``Generative self-training for cross-domain unsupervised
  tagged-to-cine mri synthesis,'' in \emph{International Conference on Medical
  Image Computing and Computer Assisted Intervention}.\hskip 1em plus 0.5em
  minus 0.4em\relax Springer, 2021, pp. 138--148.

\bibitem{wang2018interactive}
G.~Wang, W.~Li, M.~A. Zuluaga, R.~Pratt, P.~A. Patel, M.~Aertsen, T.~Doel,
  A.~L. David, J.~Deprest, S.~Ourselin \emph{et~al.}, ``Interactive medical
  image segmentation using deep learning with image-specific fine tuning,''
  \emph{IEEE transactions on medical imaging}, vol.~37, no.~7, pp. 1562--1573,
  2018.

\bibitem{valvano2021stop}
G.~Valvano, A.~Leo, and S.~A. Tsaftaris, ``Stop throwing away discriminators!
  re-using adversaries for test-time training,'' in \emph{Domain Adaptation and
  Representation Transfer, and Affordable Healthcare and AI for Resource
  Diverse Global Health}.\hskip 1em plus 0.5em minus 0.4em\relax Springer,
  2021, pp. 68--78.

\bibitem{venkataramani2018continual}
R.~Venkataramani, H.~Ravishankar, and S.~Anamandra, ``Towards continuous domain
  adaptation for healthcare,'' \emph{Neurips Workshops}, 2018.

\bibitem{goyal2017step_decay}
P.~Goyal, P.~Doll{\'a}r, R.~Girshick, P.~Noordhuis, L.~Wesolowski, A.~Kyrola,
  A.~Tulloch, Y.~Jia, and K.~He, ``Accurate, large minibatch sgd: Training
  imagenet in 1 hour,'' \emph{arXiv preprint arXiv:1706.02677}, 2017.

\bibitem{loshchilov2016anneal}
I.~Loshchilov and F.~Hutter, ``Sgdr: Stochastic gradient descent with warm
  restarts,'' \emph{arXiv preprint arXiv:1608.03983}, 2016.

\bibitem{maclaurin2015hyper_gradient}
D.~Maclaurin, D.~Duvenaud, and R.~Adams, ``Gradient-based hyperparameter
  optimization through reversible learning,'' in \emph{International conference
  on machine learning}.\hskip 1em plus 0.5em minus 0.4em\relax PMLR, 2015, pp.
  2113--2122.

\bibitem{baydin2018online_hypergradient}
A.~G. Baydin, R.~Cornish, D.~M. Rubio, M.~Schmidt, and F.~Wood, ``Online
  learning rate adaptation with hypergradient descent,'' in \emph{Sixth
  International Conference on Learning Representations (ICLR), Vancouver,
  Canada, April 30 -- May 3, 2018}, 2018.

\bibitem{dda}
S.~Li, J.~Zhang, W.~Ma, C.~H. Liu \emph{et~al.}, ``Dynamic domain adaptation
  for efficient inference,'' in \emph{CVPR}, 2021, pp. 7832--7841.

\bibitem{lee2021unsupervised}
W.~Lee, S.-Y. Byun, J.~Kim, M.~Park, and K.~Chechil, ``Unsupervised model drift
  estimation with batch normalization statistics for dataset shift detection
  and model selection,'' \emph{arXiv preprint arXiv:2107.00191}, 2021.

\bibitem{he2019deep}
Y.~He, A.~Carass, Y.~Liu, B.~M. Jedynak, S.~D. Solomon, S.~Saidha, P.~A.
  Calabresi, and J.~L. Prince, ``Deep learning based topology guaranteed
  surface and mme segmentation of multiple sclerosis subjects from retinal
  oct,'' \emph{Biomedical optics express}, vol.~10, no.~10, pp. 5042--5058,
  2019.

\bibitem{sagawa2021camelyon}
S.~Sagawa, P.~W. Koh, T.~Lee, I.~Gao, S.~M. Xie, K.~Shen, A.~Kumar, W.~Hu,
  M.~Yasunaga \emph{et~al.}, ``Extending the wilds benchmark for unsupervised
  adaptation,'' \emph{arXiv preprint arXiv:2112.05090}, 2021.

\bibitem{huang2017densely}
G.~Huang, Z.~Liu, L.~Van Der~Maaten, and K.~Q. Weinberger, ``Densely connected
  convolutional networks,'' in \emph{CVPR}, 2017, pp. 4700--4708.

\bibitem{cciccek20163d_unet}
{\"O}.~{\c{C}}i{\c{c}}ek, A.~Abdulkadir, S.~S. Lienkamp, T.~Brox, and
  O.~Ronneberger, ``3d u-net: learning dense volumetric segmentation from
  sparse annotation,'' in \emph{International Conference on Medical Image
  Computing and Computer Assisted Intervention}.\hskip 1em plus 0.5em minus
  0.4em\relax Springer, 2016, pp. 424--432.

\bibitem{paszke2019pytorch}
A.~Paszke, S.~Gross, F.~Massa, A.~Lerer, J.~Bradbury, G.~Chanan, T.~Killeen,
  Z.~Lin, N.~Gimelshein, L.~Antiga \emph{et~al.}, ``Pytorch: An imperative
  style, high-performance deep learning library,'' \emph{Advances in neural
  information processing systems}, vol.~32, 2019.

\bibitem{gatys2016image}
L.~A. Gatys, A.~S. Ecker, and M.~Bethge, ``Image style transfer using
  convolutional neural networks,'' in \emph{CVPR}, 2016, pp. 2414--2423.

\bibitem{zhu2017cycgan}
J.-Y. Zhu, T.~Park, P.~Isola, and A.~A. Efros, ``Unpaired image-to-image
  translation using cycle-consistent adversarial networks,'' in
  \emph{Proceedings of the IEEE international conference on computer vision},
  2017, pp. 2223--2232.

\bibitem{Tsai2018udas}
Y.-H. Tsai, W.-C. Hung, S.~Schulter, K.~Sohn, M.-H. Yang, and M.~Chandraker,
  ``Learning to adapt structured output space for semantic segmentation,'' in
  \emph{Proceedings of the IEEE Conference on Computer Vision and Pattern
  Recognition (CVPR)}, June 2018.

\bibitem{he2019retinal}
Y.~He, A.~Carass, S.~D. Solomon \emph{et~al.}, ``Retinal layer parcellation of
  optical coherence tomography images: Data resource for multiple sclerosis and
  healthy controls,'' \emph{Data in brief}, vol.~22, 2019.

\bibitem{koh2021wilds}
P.~W. Koh, S.~Sagawa, S.~M. Xie \emph{et~al.}, ``Wilds: A benchmark of
  in-the-wild distribution shifts,'' in \emph{ICML}.\hskip 1em plus 0.5em minus
  0.4em\relax PMLR, 2021, pp. 5637--5664.

\bibitem{bandi2018detection}
P.~Bandi, O.~Geessink, Q.~Manson, M.~Van~Dijk, M.~Balkenhol \emph{et~al.},
  ``From detection of individual metastases to classification of lymph node
  status at the patient level: the camelyon17 challenge,'' \emph{IEEE
  transactions on medical imaging}, vol.~38, no.~2, pp. 550--560, 2018.

\bibitem{litjens2018camelyon}
G.~Litjens, P.~Bandi, B.~Ehteshami~Bejnordi, O.~Geessink \emph{et~al.}, ``1399
  h\&e-stained sentinel lymph node sections of breast cancer patients: the
  camelyon dataset,'' \emph{GigaScience}, vol.~7, no.~6, p. giy065, 2018.

\bibitem{liu2020shape}
Q.~Liu, Q.~Dou, and P.-A. Heng, ``Shape-aware meta-learning for generalizing
  prostate mri segmentation to unseen domains,'' in \emph{International
  Conference on Medical Image Computing and Computer Assisted
  Intervention}.\hskip 1em plus 0.5em minus 0.4em\relax Springer, 2020, pp.
  475--485.

\bibitem{nciisbi2013}
B.~Nicholas, M.~Anant, H.~Henkjan, F.~John, K.~Justin \emph{et~al.},
  ``Nci-proc. ieee-isbi conf. 2013 challenge: Automated segmentation of
  prostate structures,'' The Cancer Imaging Archive, 2015.

\bibitem{litjens2014evaluation_prostate}
G.~Litjens, R.~Toth, W.~van~de Ven, C.~Hoeks, S.~Kerkstra, B.~van Ginneken,
  G.~Vincent, G.~Guillard, N.~Birbeck, J.~Zhang \emph{et~al.}, ``Evaluation of
  prostate segmentation algorithms for mri: the promise12 challenge,''
  \emph{Medical image analysis}, vol.~18, no.~2, pp. 359--373, 2014.

\bibitem{lemaitre2015prostate}
G.~Lema{\^\i}tre, R.~Mart{\'\i}, J.~Freixenet, J.~C. Vilanova, P.~M. Walker,
  and F.~Meriaudeau, ``Computer-aided detection and diagnosis for prostate
  cancer based on mono and multi-parametric mri: a review,'' \emph{Computers in
  biology and medicine}, vol.~60, pp. 8--31, 2015.

\end{thebibliography}
 \bibliographystyle{IEEEtran}
   
\end{document}